\documentclass[sigconf]{acmart}

\usepackage{graphicx}
\usepackage{balance}  

\usepackage{graphics}
\usepackage{xspace}
\usepackage[ruled, vlined, linesnumbered]{algorithm2e} 

\usepackage{booktabs}
\usepackage{soul}
\usepackage{balance}

\usepackage{float}
\usepackage{subcaption}

\usepackage{algorithmicx, algpseudocode}
\usepackage{amsmath}
\usepackage{amsfonts}
\usepackage{amsbsy}
\usepackage{amsthm}
\usepackage[]{caption}
\usepackage{multirow}
\usepackage[mathscr]{eucal} 
\usepackage{mathtools}
\usepackage{verbatim}
\usepackage{tabularx}
\usepackage{enumitem}
\usepackage{array}
\usepackage{soul}
\usepackage{dsfont}
\usepackage{url}
\usepackage{bbm}
\usepackage{pifont}
\usepackage{xcolor,colortbl}
\usepackage[inkscapeformat=png]{svg}
\usepackage{makecell}
\usepackage{contour}
\usepackage{ulem}
\contourlength{0.8pt}

\newcommand{\myuline}[1]{%
  \uline{\phantom{#1}}%
  \llap{\contour{white}{#1}}%
}

\usepackage{xspace}

\definecolor{newgreen}{rgb}{0.0, 0.5, 0.0}
\definecolor{newred}{rgb}{0.81,0.1,0.26}
\definecolor{babyblue}{rgb}{0.54, 0.81, 0.94}
\definecolor{negcol}{rgb}{0.96, 0.76, 0.76}
\definecolor{poscol}{rgb}{0.63, 0.79, 0.95}
\definecolor{sunwoogreen}{rgb}{0.66, 0.89, 0.63}
\definecolor{sunwoogreentwo}{rgb}{0.53, 0.81, 0.98}
\definecolor{dodgerblue}{rgb}{0.12, 0.56, 1.0}
\definecolor{crimson}{rgb}{0.86, 0.08, 0.24}
\definecolor{limegreen}{rgb}{0.2, 0.8, 0.2}
\definecolor{backredcolor}{rgb}{0.91, 0.45, 0.32}

\let\oldnl\nl
\newcommand{\nlnonumber}{\renewcommand{\nl}{\let\nl\oldnl}}

\AtBeginDocument{%
  \providecommand\BibTeX{{%
    \normalfont B\kern-0.5em{\scshape i\kern-0.25em b}\kern-0.8em\TeX}}}

\copyrightyear{2024}
\acmYear{2024}
\setcopyright{acmlicensed}\acmConference[KDD '24]{Proceedings of the 30th ACM SIGKDD Conference on Knowledge Discovery and Data Mining}{August 25--29, 2024}{Barcelona, Spain}
\acmBooktitle{Proceedings of the 30th ACM SIGKDD Conference on Knowledge Discovery and Data Mining (KDD '24), August 25--29, 2024, Barcelona, Spain}
\acmDOI{10.1145/3637528.3671845} 
\acmISBN{979-8-4007-0490-1/24/08}  

\begin{document}

    \title{SLADE: Detecting Dynamic Anomalies in Edge Streams 
     without Labels via Self-Supervised Learning}
    
	
	\author{Jongha Lee}
        \orcid{0000-0001-7197-3529}
	\affiliation{%
    	\institution{KAIST}
            \city{Seoul}
            \country{Republic of Korea}
	}
	\email{jhsk777@kaist.ac.kr}

        \author{Sunwoo Kim}
        \orcid{0009-0006-6002-169X}
	\affiliation{%
		\institution{KAIST}
            \city{Seoul}
            \country{Republic of Korea}
	}
	\email{kswoo97@kaist.ac.kr}
	
	\author{Kijung Shin}
        \orcid{0000-0002-2872-1526}
	\affiliation{%
		\institution{KAIST}
            \city{Seoul}
            \country{Republic of Korea}
	}
	\email{kijungs@kaist.ac.kr}

    \begin{abstract}
		To detect anomalies in real-world graphs, such as social, email, and financial networks, various approaches have been developed.
While they typically assume static input graphs, most real-world graphs grow over time, naturally represented as edge streams. 
In this context, we aim to achieve three goals: (a) instantly detecting anomalies as they occur, (b) adapting to dynamically changing states, and (c) handling the scarcity of dynamic anomaly labels.

In this paper, we propose \method (\underline{S}elf-supervised \underline{L}earning for \underline{A}nomaly \underline{D}etection in \underline{E}dge Streams) for rapid detection of dynamic anomalies in edge streams, without relying on labels.
\method detects the shifts of nodes into abnormal states by observing deviations in their interaction patterns over time. 
To this end, it trains a deep neural network to perform two self-supervised tasks: (a) minimizing drift in node representations and (b) generating long-term interaction patterns from short-term ones. Failure in these tasks for a node signals its deviation from the norm.
Notably, the neural network and tasks are carefully designed so that all required operations can be performed in constant time (w.r.t. the graph size) in response to each new edge in the input stream. In dynamic anomaly detection across four real-world datasets, \method outperforms nine competing methods, even those leveraging label supervision.
Our code and datasets are available at \repo.
	\end{abstract}
	
	\newcommand\red[1]{\textcolor{red}{#1}}
\newcommand\teal[1]{\textcolor{teal}{#1}}
\newcommand\blue[1]{\textcolor{blue}{#1}}
\newcommand\gray[1]{\textcolor{gray}{#1}}

\definecolor{peace}{RGB}{228, 26, 28}
\definecolor{love}{RGB}{55, 126, 184}
\definecolor{joy}{RGB}{77, 175, 74}
\definecolor{kindness}{RGB}{152, 78, 163}

\newcommand\peace[1]{\textcolor{peace}{#1}}
\newcommand\love[1]{\textcolor{love}{#1}}
\newcommand\joy[1]{\textcolor{joy}{#1}}
\newcommand\kindness[1]{\textcolor{kindness}{#1}}

\newcommand\kijung[1]{\textcolor{peace}{[Kijung: #1]}}
\newcommand\fanchen[1]{\textcolor{love}{[Fanchen: #1]}}
\newcommand\minyoung[1]{\textcolor{joy}{[Minyoung: #1]}}
\newcommand\sunwoo[1]{\textcolor{kindness}{[Sunwoo: #1]}}

\newcommand{\change}{\textcolor{red}}
\newcommand{\done}{\red{[DONE]}\xspace}
\newcommand{\todo}[1]{\red{[TODO: #1]}\xspace}
\newcommand{\ongoing}{\red{(ONGOING)}\xspace}
\newcommand{\tb}[1]{\textbf{#1}\xspace}

\newcommand{\smallsection}[1]{{\vspace{0.02in} \noindent {{\myuline{\smash{\bf #1:}}}}}}
\newtheorem{obs}{\textbf{Observation}}
\newtheorem{defn}{\textbf{Definition}}
\newtheorem{thm}{\textbf{Theorem}}
\newtheorem{axm}{\textbf{Axiom}}
\newtheorem{lma}{\textbf{Lemma}}
\newtheorem{cor}{\textbf{Corollary}}
\newtheorem{problem}{\textbf{Problem}}
\newtheorem{pro}{\textbf{Problem}}
\newtheorem{remark}{\textbf{Remark}}

\newcommand\und[1]{\underline{#1}}

\newcommand{\calA}{\mathcal{A}}
\newcommand{\calB}{\mathcal{B}}
\newcommand{\calC}{\mathcal{C}}
\newcommand{\calD}{\mathcal{D}}
\newcommand{\calE}{\mathcal{E}}
\newcommand{\calF}{\mathcal{F}}
\newcommand{\calG}{\mathcal{G}}
\newcommand{\calH}{\mathcal{H}}
\newcommand{\calI}{\mathcal{I}}
\newcommand{\calJ}{\mathcal{J}}
\newcommand{\calK}{\mathcal{K}}
\newcommand{\calL}{\mathcal{L}}
\newcommand{\calM}{\mathcal{M}}
\newcommand{\calN}{\mathcal{N}}
\newcommand{\calO}{\mathcal{O}}
\newcommand{\calP}{\mathcal{P}}
\newcommand{\calQ}{\mathcal{Q}}
\newcommand{\calR}{\mathcal{R}}
\newcommand{\calS}{\mathcal{S}}
\newcommand{\calT}{\mathcal{T}}
\newcommand{\calU}{\mathcal{U}}
\newcommand{\calV}{\mathcal{V}}
\newcommand{\calW}{\mathcal{W}}
\newcommand{\calX}{\mathcal{X}}
\newcommand{\calY}{\mathcal{Y}}
\newcommand{\calZ}{\mathcal{Z}}
\newcommand{\weightg}{\omega_{g}}
\newcommand{\weightm}{\omega_{m}}
\newcommand{\repo}{\url{https://github.com/jhsk777/SLADE}}

\newcommand{\veca}{\boldsymbol{a}}
\newcommand{\vecb}{\boldsymbol{b}}
\newcommand{\vecc}{\boldsymbol{c}}
\newcommand{\vecd}{\boldsymbol{d}}
\newcommand{\vece}{\boldsymbol{e}}
\newcommand{\vecf}{\boldsymbol{f}}
\newcommand{\vecg}{\boldsymbol{g}}
\newcommand{\vech}{\boldsymbol{h}}
\newcommand{\veci}{\boldsymbol{i}}
\newcommand{\vecj}{\boldsymbol{j}}
\newcommand{\veck}{\boldsymbol{k}}
\newcommand{\vecl}{\boldsymbol{l}}
\newcommand{\vecm}{\boldsymbol{m}}
\newcommand{\vecn}{\boldsymbol{n}}
\newcommand{\veco}{\boldsymbol{o}}
\newcommand{\vecp}{\boldsymbol{p}}
\newcommand{\vecq}{\boldsymbol{q}}
\newcommand{\vecr}{\boldsymbol{r}}
\newcommand{\vecs}{\boldsymbol{s}}
\newcommand{\vect}{\boldsymbol{t}}
\newcommand{\vecu}{\boldsymbol{u}}
\newcommand{\vecv}{\boldsymbol{v}}
\newcommand{\vecw}{\boldsymbol{w}}
\newcommand{\vecy}{\boldsymbol{y}}
\newcommand{\vecz}{\boldsymbol{z}}

\newcommand{\vecsp}{\boldsymbol{s}'}

\newcommand{\matA}{\mathbf{A}}
\newcommand{\matB}{\mathbf{B}}
\newcommand{\matC}{\mathbf{C}}
\newcommand{\matD}{\mathbf{D}}
\newcommand{\matE}{\mathbf{E}}
\newcommand{\matF}{\mathbf{F}}
\newcommand{\matG}{\mathbf{G}}
\newcommand{\matH}{\mathbf{H}}
\newcommand{\matI}{\mathbf{I}}
\newcommand{\matJ}{\mathbf{J}}
\newcommand{\matK}{\mathbf{K}}
\newcommand{\matL}{\mathbf{L}}
\newcommand{\matM}{\mathbf{M}}
\newcommand{\matN}{\mathbf{N}}
\newcommand{\matO}{\mathbf{O}}
\newcommand{\matP}{\mathbf{P}}
\newcommand{\matQ}{\mathbf{Q}}
\newcommand{\matR}{\mathbf{R}}
\newcommand{\matS}{\mathbf{S}}
\newcommand{\matT}{\mathbf{T}}
\newcommand{\matU}{\mathbf{U}}
\newcommand{\matV}{\mathbf{V}}
\newcommand{\matW}{\mathbf{W}}
\newcommand{\matX}{\mathbf{X}}
\newcommand{\matY}{\mathbf{Y}}
\newcommand{\matZ}{\mathbf{Z}}

\newcommand{\method}{\textsc{SLADE}\xspace}

\newcommand{\SedanSpot}{SedanSpot\xspace}
\newcommand{\MIDAS}{MIDAS\xspace}
\newcommand{\FFADE}{F-FADE\xspace}
\newcommand{\Anoedgel}{Anoedge-l\xspace}
\newcommand{\JODIE}{JODIE\xspace}
\newcommand{\Dyrep}{Dyrep\xspace}
\newcommand{\TGAT}{TGAT\xspace}
\newcommand{\TGN}{TGN\xspace}
\newcommand{\SAD}{SAD\xspace}
\newcommand{\MLP}{MLP\xspace}
\newcommand{\GCN}{GCN\xspace}
\newcommand{\GAT}{GAT\xspace}

\newcommand{\attention}{\textsc{ATT}\xspace}
\newcommand{\aggregation}{\textsc{AGG}\xspace}
\newcommand{\MultiheadAtt}{\textsc{MAB}\xspace}
\newcommand{\WithinAttentionFull}{within attention\xspace}
\newcommand{\WithinAttention}{\textsc{WithinATT}\xspace}
\newcommand{\PE}{\textsc{WithinOrderPE}\xspace}

\newcommand{\methodgrid}{\textsc{MiDaS-Grid}\xspace}
\newcommand{\methodauto}{\textsc{MiDaS}\xspace}

\definecolor{myred}{RGB}{195, 79, 82}
\definecolor{mygreen}{RGB}{86, 167 104}
\definecolor{myblue}{RGB}{74, 113 175}

\newcommand{\bigcell}[2]{\begin{tabular}{@{}#1@{}}#2\end{tabular}}

\let\oldnl\nl
\newcommand{\nonl}{\renewcommand{\nl}{\let\nl\oldnl}}

	\maketitle

	\section{Introduction}
\label{sec:intro}

The evolution of web technologies has dramatically enhanced human life. 
Platforms such as email and social networks have made people communicating with diverse individuals and accessing helpful information easier. 
Additionally, e-commerce has enabled people to engage in economic activities easily. However, as convenience has increased, many problems have emerged, such as financial crimes, social media account theft, and spammer that exploited it.

Many graph anomaly detection techniques ~\citep{pourhabibi2020fraud, ma2021comprehensive} have been developed for tackling these problems.
These techniques involve representing the interactions between users as a graph, thereby harnessing the connectivity between users to effectively identify anomalies. 
However, graph anomaly detection in real-world scenarios poses several challenges, as discussed below.

\smallsection{C1) Time Delay in Detection}
While most graph anomaly detection methods assume static input graphs,  real-world graphs evolve over time with continuous interaction events.
In response to continuous interaction events, it is important to quickly identify anomalies. 
Delaying the detection of such anomalies can lead to increasing harm to benign nodes as time passes.
However, employing static graph-based methods repeatedly on the entire graph, whenever an interaction event occurs, inevitably leads to significant delays due to the substantial computational expenses involved.
To mitigate delays, it is necessary to model continuous interaction events as edge streams and employ incremental computation to assess the abnormality of each newly arriving edge with detection time constant regardless of the accumulated data size.

Many studies~\citep{eswaran2018sedanspot,bhatia2020midas} have developed anomaly detection methods for edge streams, leveraging incremental computation. 
However, as these methods are designed to target specific anomaly types (e.g., burstiness), lacking learning-based components,  they are often limited in capturing complex ones deviating from targeted types.

\smallsection{C2) Dynamically Changing States}
In web services, users can exhibit dynamic states varying over time.
That is, a user's behavior can be normal during one time period but abnormal during another.
For example, a normal user's account can be compromised and then manipulated to disseminate promotional messages. As a result, the user's state transitions from normal to abnormal. 
Such a user can be referred to as a dynamic anomaly, and detecting dynamic anomalies presents a greater challenge compared to the relatively easier task of identifying static anomalies.

Addressing this challenge can be facilitated by tracking the evolution of node characteristics over time, and to this end,  dynamic node representation learning methods~\citep{xu2020inductive,tgn_icml_grl2020} can be employed.
However, existing dynamic node representation learning methods~ require label information for the purpose of anomaly detection, which is typically scarce, as elaborated in the following paragraph.

\smallsection{C3) Lack of Anomaly Labels}
Deep neural networks, such as graph neural networks, 
have proven effective in detecting complex anomalies within graph-structured data.
However, they typically rely on label supervision for training, which can be challenging to obtain, especially for dynamic anomalies.
Assuming the absence of anomaly labels, various unsupervised training techniques~\citep{ding2019deep,liu2021cola} have been developed for anomaly detection in static graphs.

However, to the best of our knowledge, existing methods cannot simultaneously address all three challenges.
They either lack the ability to detect dynamic anomalies, lack incremental detection suitable for edge streams, or rely on labels.

In this work, we propose \method (\underline{S}elf-supervised \underline{L}earning for \underline{A}nomaly \underline{D}etection in \underline{E}dge Streams) to simultaneously tackle all three challenges.
Its objective is to incrementally identify dynamic anomalies in edge streams, without relying on any label supervision.
To achieve this, \method learns dynamic representations of nodes over time, which capture their long-term interaction patterns, by training a deep neural network to perform two self-supervised tasks: (a) minimizing drift in the representations and (b) generating long-term interaction patterns from short-term ones.
Poor performance on these tasks for a node indicates its deviation from its interaction pattern, signaling a potential transition to an abnormal state. Notably, our careful design ensures that all required operations are executed in constant time (w.r.t. the graph size) in response to each new edge in the input stream.
Our experiments involving dynamic anomaly detection by 9 competing methods across 4 real-world datasets confirm the strengths of \method, outlined below:
\begin{itemize}[leftmargin=*]
    \item \textbf{Unsupervised:} We propose \method to detect complex dynamic anomalies in edge streams, without relying on label supervision.
    \item \textbf{Effective:} 
    In dynamic anomaly detection, \method shows an average improvement of 12.80\% and 4.23\% (in terms of AUC) compared to the best-performing (in each dataset) unsupervised and supervised competitors, respectively.
    \item \textbf{Constant Inference Speed:} We show both theoretically and empirically that, once trained, \method requires a constant amount time per edge for dynamic anomaly detection in edge streams. 
\end{itemize}

	\section{Related Work}
\label{sec:related}

In this section, we provide a concise review of graph-based anomaly detection methods, categorized based on the nature of input graphs.

\subsection{Anomaly Detection in CTDGs}
\label{sec:anomaly_CTDG}
A \textit{continuous-time dynamic graph} (CTDG) is a stream of edges accompanied by timestamps, which we also term an \textit{edge stream}.

\smallsection{Unsupervised Anomaly Detection in CTDGs}
The majority of unsupervised anomaly detection techniques applied to CTDGs aim to identify specific anomaly types.
Sedanspot~\citep{eswaran2018sedanspot} focuses on detecting (a) bursts of activities and (b) bridge edges between sparsely connected parts of the input graph.
MIDAS~\citep{bhatia2020midas} aims to spot bursts of interactions within specific groups, and F-FADE~\citep{chang2021f} is effective for identifying
sudden surges in interactions between specific pairs of nodes and swift changes in the community memberships of nodes.
Lastly,
AnoGraph~\citep{bhatia2021sketch} spots dense subgraph structures along with the anomalous edges contained within them. 
These methods are generally efficient, leveraging incremental computation techniques. 
Nonetheless, as previously mentioned, many of these approaches lack learnable components and thus may encounter challenges in identifying complex anomaly patterns, i.e., deviations from normal patterns in various aspects that may not be predefined.

\smallsection{Representation Learning in CTDGs}
Representation learning in CTDGs involves maintaining and updating representations of nodes in response to each newly arriving edge, capturing evolving patterns of nodes over time.
To this end, several neural-network architectures have been proposed.
JODIE~\citep{kumar2019predicting} utilizes recurrent-neural-network (RNN) modules to obtain dynamic {node} representations. 
Dyrep~\citep{trivedi2019dyrep} combines  
a deep temporal point process \citep{aalen2008survival} with RNNs.
TGAT~\citep{xu2020inductive} leverages temporal encoding and graph attention~\citep{velickovic2017graph} to incorporate temporal information during neighborhood aggregation. 
Based on temporal smoothness, DDGCL~\citep{tian2021self} learns dynamic node representations by contrasting those
of the same nodes in two nearby temporal views.
TGN~\citep{tgn_icml_grl2020} utilizes a memory module to capture and store long-term patterns. This module is updated using an RNN for each node, providing representations that encompass both temporal and spatial characteristics.
Many subsequent studies~\citep{wang2021apan,wang2021adaptive} have also adopted memory modules. Assuming a gradual process where memories for nodes do not show substantial disparities before and after updates, DGTCC~\citep{huang2022dynamic} contrasts memories before and after updates for training.

For the purpose of anomaly detection,
dynamic node representations can naturally be used as inputs for a classifier, which is trained in a supervised manner using anomaly labels.
Recently, more advanced anomaly detection methods based on representation learning in CTDGs have been proposed. 
SAD~\citep{tian2023sad} combines a memory bank with pseudo-label contrastive learning, which, however, requires anomaly labels for training.

\subsection{Anomaly Detection in Other Graph Models}
\label{sec:related:others}
In this subsection, we introduce anomaly detection approaches applied to other graph models.

\smallsection{Anomaly Detection in Static Graphs}
Many approaches~\citep{li2017radar,ding2019deep,ding2021few} have been developed for anomaly detection in a \textit{static graph}, which does not contain any temporal information.
Among them, we focus on those leveraging graph self-supervised learning, which effectively deals with the absence of anomaly labels.
ANEMONE~\citep{jin2021anemone} contrasts node representations obtained from (a) features alone and (b) both graph topology and features, identifying nodes with substantial differences as anomalies.
DOMINANT~\citep{ding2019deep} aims to reconstruct graph topology and attributes, using a graph-autoencoder module, and anomalies are identified based on reconstruction error.
These approaches assume a static graph, and their extensions to dynamic graphs are not trivial, as node representations need to evolve over time to accommodate temporal changes.

\smallsection{Anomaly Detection in DTDGs}
 A \textit{discrete-time dynamic graph} (DTDG) corresponds to a sequence of graphs occurring at each time instance, also referred to as a \textit{graph stream}.
 Formally, a DTDG is a set $\left\{\mathcal{G}^{1}, \mathcal{G}^{2},\cdots, \mathcal{G}^{T}\right\}$ where $\mathcal{G}^{t}=\left\{\mathcal{V}^{t},\mathcal{E}^{t} \right\}$
 is the graph snapshot at time $t$, and 
 $\mathcal{V}^{t}$ and $\mathcal{E}^{t}$ are node- and edge-set in $\mathcal{G}^{t}$, respectively \citep{kazemi2020representation}.
Several methods have been developed for detecting anomalies, with an emphasis on anomalous edges, in a DTDG, which is a sequence of graphs at each time instance.
Netwalk~\citep{yu2018netwalk} employs random walks and autoencoders to create similar representations for nodes that frequently interact with each other. Then, it identifies interactions between nodes with distinct representations as anomalous. 
AddGraph~\citep{zheng2019addgraph} constructs node representations through an attentive combination of (a) short-term structural patterns within the current graph (and a few temporally-adjacent graphs) captured by a graph neural network (spec., GCN~\citep{kipf2016semi}) and (b) long-term patterns captured by an RNN (spec., GRU~\citep{cho2014learning}).
These node representations are then employed to evaluate the anomalousness of edges.
Instead, transformers~\citep{vaswani2017attention} are employed in TADDY~\citep{liu2021anomaly} to acquire node representations that capture both global and local structural patterns.
Note that these methods designed for DTDGs are less suitable for time-critical applications when compared to those for designed CTDGs, as discussed in Section~\ref{sec:prelim}.
Moreover, technically, these methods are trained to distinguish edges and non-edge node pairs, while our proposed method  contrasts long-term and short-term patterns for training.

\section{Problem Description}
\label{sec:prelim}

In this section, we introduce notations and, based on them,  define the problem of interest, dynamic node anomaly detection in CTDGs.

\smallsection{Notations}
A continuous-time dynamic graph (CTDG) $\mathcal{G}=(\delta_{1},\delta_{2},$ $\cdots)$ is a stream (i.e., continuous sequence) of temporal edges with timestamps. Each temporal edge $\delta_{n}=(v_{i},v_{j},t_n)$ arriving at time $t_n$ is directional from the \textit{source node} $v_i$ to the \textit{destination node} $v_j$. The temporal edges are ordered chronologically, i.e., $t_n\leq t_{n+1}$ holds for each $n \in \{1,2,\cdots\}$.
We denote by $\mathcal{V}(t)=\bigcup_{(v_{i},v_{j},t_n) \in \mathcal{G} \wedge t_n\leq t}\{v_{i},v_{j}\}$ the temporal set of nodes arriving at time $t$ or earlier.

\smallsection{Problem Description}
We consider a CTDG $\mathcal{G}=(\delta_{1},\delta_{2},\cdots)$, where each temporal edge $\delta_{n}=(v_{i},v_{j},t_{n})$ indicates a behavior of the source node $v_i$ towards the destination node $v_j$ at time $t_{n}$.
That is, the source node represents the ``actor'' node.
We aim to accurately classify the current dynamic status of each node, which is either \textit{normal} or \textit{abnormal}.
We address this problem in an unsupervised setting. That is, we do not have access to the dynamic states of any nodes at any time as input.
In our experimental setups, the ground-truth dynamic states are used only for evaluation purposes.

\smallsection{Real-world Scenarios}
Due to its substantial impact, this problem has been explored in many previous studies~\citep{kumar2019predicting,trivedi2019dyrep,xu2020inductive,tgn_icml_grl2020}, but in supervised settings. 
For instance, in web services, a normal user's account can be compromised and then exploited to circulate promotional messages. In such a case, the user's state transitions from normal to abnormal. 
Detecting such transitions promptly is crucial to minimize the inconvenience caused by the dissemination of promotional messages.

\smallsection{Why CTDGs?}
Anomaly detection methods designed for DTDGs (refer to Section~\ref{sec:related:others}) process input on a per-graph basis, where edges need to be aggregated over time to form a graph. Thus, they are susceptible to delays in predicting the current state of nodes upon the arrival of an edge. 
In theory, methods designed for static graphs can be applied to our problem by re-running them whenever a new edge arrives. However, this straightforward application makes their time complexity per edge (super-)linear in the graph size, causing notable delays.
However, CTDG-based methods, including our proposed one, process the arriving edge, whenever it arrives, typically with constant time complexity regardless of the accumulated graph size.
As discussed in Section~\ref{sec:intro},
this highlights the advantages of CTDG-based methods in time-critical applications, including (dynamic) anomaly detection.

\smallsection{Comparison with Edge Anomaly Detection}
In many scenarios, the dynamic state of an actor node can be equated with the anomalousness (or maliciousness) of its behavior.
In other words, an actor node is assumed to be in the abnormal state if and only if it performs anomaly (or malicious) behavior.
In such cases, this task shares some similarities with detecting anomalous edges. However, it should be noticed that the predicted current node states are used for predicting the anomalousness of future interactions (see Section~\ref{sec:method:anomalyscore})
rather than assessing the anomalousness of interactions that have already occurred.
Hence, for effective dynamic node anomaly detection, it is important to consider node-wise behavioral dynamics over time.

\begin{figure*}[t]
    \centering
    \includegraphics[width=1\linewidth]{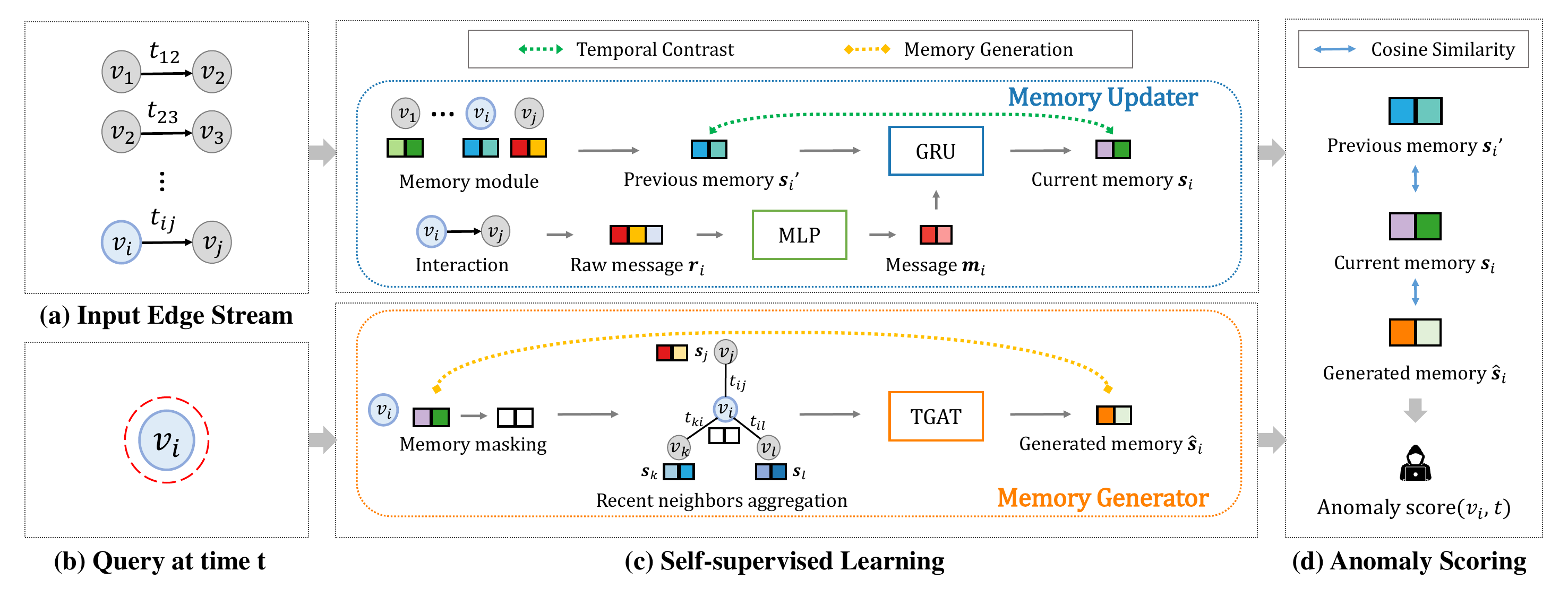}
    \centering
    \caption{ \label{fig:diagram}
        Overview of \method, whose objective is to measure the anomaly score of a query node at any time.
        For each newly arriving edge, \method updates the memory vector of each endpoint using GRU.
        Given a query node, \method masks the memory vector of the node and approximately regenerates it based on its recent interactions using TGAT. Then, it measures the anomaly score of the query node based on the similarities (1) between previous and current memory vectors (related to \textbf{{S1}}) and (2) between current and generated memory vectors (related to \textbf{{S2}}).
        \method aims to maximize these similarities for model training.
    }
\end{figure*}

\section{Proposed Method: \method}
\label{sec:method}

In this section, we present \method (\underline{S}elf-supervised \underline{L}earning for \underline{A}nomaly \underline{D}etection in \underline{E}dge Streams), our proposed method for unsupervised dynamic anomaly detection in CTDGs.

The underlying intuition is that nodes in the normal state tend to exhibit structurally and temporally similar interaction patterns over time~\citep{benson2018sequences, anderson2014dynamics}, while those in the abnormal state do not because repeating similar abnormal actions increases the risk of detection.
Motivated by this idea, we consider two key assumptions:
\begin{itemize}[leftmargin=*]
    \item \textbf{A1. Stable Long-Term Interaction Patterns:} Nodes in the normal state tend to repetitively engage in similar interactions over a long-term period.
    This stable long-term interaction pattern exhibits minimal variation within short-time intervals.
    \item \textbf{A2. Potential for Restoration of Patterns:} It would be feasible to accurately regenerate the long-term interaction patterns of the nodes in the normal state using recent interaction information. 
\end{itemize}
They account for both structural and temporal aspects of normal nodes. While \textbf{A1} focuses on temporal aspects, \textbf{A2} takes a further step by specifying the extent of structural similarities over time.

Upon these assumptions, \method employs two self-supervised tasks for training its model (i.e., deep neural network) for maintaining and updating a dynamic representation of each node, which we expect to capture its long-term interaction pattern.
\begin{itemize}[leftmargin=*]
    \item \textbf{S1. Temporal Contrast:} 
    This aims to minimize drift in dynamic node representations over short-term periods (related to \textbf{A1}).
    \item \textbf{S2. Memory Generation:} This aims to accurately generate dynamic node representations based only on recent interactions (related to \textbf{A2}).
\end{itemize}
By being trained for \textbf{S1} and \textbf{S2}, the model is expected to learn normal interaction patterns satisfying \textbf{A1} and \textbf{A2}.
Once the model is trained, \method identifies nodes for which the model performs poorly on \textbf{S1} and \textbf{S2}, as these nodes potentially deviate from presumed normal interaction patterns.

Specifically, to obtain dynamic representations, \method employs neural networks in combination with memory modules (see Section~\ref{sec:method:proposedmodule}).
The memories (i.e, stored information), which reflect the normal patterns of nodes, are updated and regenerated to minimize our self-supervised losses related to \textbf{S1} and \textbf{S2}
(see Section~\ref{sec:method:proposed}).
Lastly, these dynamic representations are compared with their past values and momentary representations to compute anomaly scores for nodes (see Section~\ref{sec:method:anomalyscore}).
The overview of \method is visually presented in Figure~\ref{fig:diagram}, and the following subsections provide detailed descriptions of each of its components.

\subsection{Core Modules of \method}
\label{sec:method:proposedmodule}
In order to incrementally compute the dynamic representation of each node, \method employs three core modules:  
\begin{itemize}[leftmargin=*]
    \item \textbf{Memory Module:} These time-evolving parameter vectors represent the long-term interaction patterns of each node, i.e., how a node's interaction has evolved over time.
    \item \textbf{Memory Updater:} This neural network captures evolving characteristics of nodes' interaction patterns.
    It is employed to update the memory (i.e., stored information).
    \item \textbf{Memory Generator:} This neural network is used to generate the memory of a target node from its recent interactions.
\end{itemize}
Below, we examine the details of each module in order.

\smallsection{Memory Module}
In \method, the dynamic representation of each node, 
which represents its long-term interaction patterns,
is stored and updated in a memory module introduced by \citet{tgn_icml_grl2020}.
Specifically,
the memory module consists of a memory vector $\vecs_{i}$ for each  node $v_{i}$, and 
each $\vecs_{i}$ captures the interactions of the node $v_{i}$ up to the current time. 
When each node $v_{i}$ first emerges in the input CTDG, $\vecs_{i}$ is initialized to a zero vector. 
As $v_{i}$ participates in interactions, $\vecs_{i}$ is continuously  updated by the memory updater, described in the following subsection.

\smallsection{Memory Updater}
Whenever a node participates in a new interaction, the memory updater gradually updates its memory vector, aiming to represent its stable long-term interaction pattern. 
First, each interaction is transformed into raw messages~\citep{tgn_icml_grl2020}.
Each raw message consists of the encoded time difference between the most recent appearance of one endpoint and the present time, along with the memory vector of the other endpoint.
For instance, upon the arrival of a temporal edge $(v_{i},v_{j},t_{ij})$ at time $t_{ij}$, with $v_{i}$ and $v_{j}$ having memory vectors $\vecsp_{i}$ and $\vecsp_{j}$, respectively, the raw messages for the source and destination nodes are created as follows:
\begin{equation}\label{eq:message function}
\vecr_{i}=[\vecsp_{j}\vert\vert \phi(t_{ij}-t_{i}^{-})], \ \ 
\vecr_{j}=[\vecsp_{i}\vert\vert \phi(t_{ij}-t_{j}^{-})], 
\end{equation}
where $\vert\vert$ denotes a concatenate operator, 
and $t_{i}^{-}$ denotes the time of the last interaction of $v_i$ before the interaction time $t_{ij}$.
Following~\citep{cong2022we}, 
as the time encoding function $\phi(\cdot)$, we use
\begin{equation}\label{eq:timeencoding}
    \phi(t')  = cos\left(t' \cdot [\alpha^{-\frac{0}{\beta}} \vert\vert  \alpha^{-\frac{1}{\beta}}\vert\vert \cdots \vert\vert \alpha^{-\frac{d_{t}-1}{\beta}}]\right), 
\end{equation}
where $\cdot$ denotes the inner product, and $d_{t}$ denotes the dimension of encoded vectors.
Scalars $\alpha$, $\beta$ and $d_{t}$ are hyperparameters.

Then, the raw message $\vecr_{i}$ is converted to a message $\vecm_{i}$, employing an MLP and then used to update from $\vecsp_{i}$ to $\vecs_{i}$, as follows:
\begin{equation}\label{eq:memory update}
\vecs_{i}=\text{GRU}(\vecm_{i},\vecsp_{i}) \text{ where } \vecm_{i}=\text{MLP}(\vecr_{i}). 
\end{equation}
Here
$\vecs_{i}$ is the memory vector for node $v_{i}$ after time $t_{ij}$, persisting until a new interaction involving node $v_{i}$ occurs
(see Section~\ref{sec:app:exp:strucutre} (RQ5) for exploration of alternatives to GRU \cite{cho2014learning}).
We also maintain $\vecsp_{i}$ (i.e., the previous value of the memory vector) for its future usage.

\smallsection{Memory Generator}
The memory generator aims to restore the memory vectors, which represent existing long-term interaction patterns, based on short-term interactions.
The generated vectors are used for training and anomaly scoring, as described later. 

The first step of generation is to (temporarily) mask the memory vector of a target node at current time $t$ to a zero vector. 
Then, the memory vectors of its at most $k$ latest (1-hop) neighbors and the corresponding time information of their latest interactions are used as inputs of an encoder.
As the encoder, we employ TGAT~\citep{xu2020inductive}, which attends more to recent interactions (see Section~\ref{sec:app:exp:strucutre} (RQ5) for exploration of alternatives).
That is, for a target node $v_i$ and the current time $t$, we generate its memory vector as follows:
\begin{multline}\label{eq:memory update}
 \hat{\vecs}_{i}=\text{MultiHeadAttention}(\vecq,\textbf{K},\textbf{V}),  \text{ where }  \ \vecq=\mathbf{\phi}(t-t),\\
  \text{and }  \textbf{K}=\textbf{V} = \left [ \mathbf{s}_{n_1}||\phi(t-t'_{n_1}),\cdots,\textbf{s}_{n_k}||\phi(t-t'_{n_k}) \right].  \nonumber
\end{multline}
Here $\{n_{1}, ..., n_{k}\}$ denote the indices of the neighbors of the target node $v_i$, $\{t'_{n_1}, ..., t'_{n_k}\}$ denote the timestamps of the most recent interactions with them before $t$, and $\hat{\textbf{s}}_{i}$ denotes the generated memory vector of $v_i$ at $t$.
As the memory vector has been masked, only time information is used for query component $\vecq$.

\subsection{Training Objective and Procedure}
\label{sec:method:proposed}

For the proposed temporal contrast task (\textbf{S1}) and memory generation task (\textbf{S2}), \method is trained to minimize two loss components:
\begin{itemize}[leftmargin = 7pt]
    \item \textbf{Temporal Contrast Loss:} It encourages the agreement between previous and updated memory vectors.
    \item \textbf{Memory Generation Loss:} It encourages the similarity between retained and generated memory vectors. 
\end{itemize}
Below, we describe each self-supervised loss component and then the entire training process in greater detail.

\smallsection{Temporal Contrast Loss}
For \textbf{S1}, we aim to minimize drift in dynamic node representations within short time intervals. 
For the memory vector $\vecs_{i}$ of each node $v_{i}$, we use the previous memory vector $\vecsp_{i}$ as the positive sample and the memory vectors of the other nodes as negative samples.
Specifically, for each interaction at time $t$ of $v_i$, we aim to minimize the following loss: 
\begin{equation}\label{eq:memory drift contrast}
\begin{split}
& \ell_{c}(v_{i},t)=-\log\frac{\exp{(sim(\vecs_{i}(t^+),\vecsp_{i}(t^+)}))}{\sum_{k=1}^{\left|\mathcal{V}(t^+) \right|}\exp{(sim(\vecs_{i}(t^+),\vecs_{k}(t^+)))}},
\end{split}
\end{equation}
where $\vecs_i(t^+)$ is the current memory vector $\vecs_i$ right after processing the interaction at $t$ (denoted as $t^+$ to distinguish it from $t$ right before processing the interaction),
$\vecs'_i(t^+)$ is the previous memory vector, 
and
$sim$ is the cosine similarity function, i.e., $sim(\vecu, \vecv) = \frac{\lVert \vecu \cdot \vecv \rVert}{\lVert \vecu \rVert \cdot \lVert \vecv \rVert}$. 

\smallsection{Memory Generation Loss}
For \textbf{S2}, we aim to accurately generate dynamic node representations from recent interactions (i.e., short-term patterns).
For each node $v_{i}$, we expect its generated memory vector $\hat{\vecs}_{i}$ to be matched well with the (temporally masked) memory vector ${\vecs}_{i}$, relative to other memory vectors.  
Specifically, for each interaction at time $t$ of $v_i$, we aim to minimize the following loss:
\begin{equation}\label{eq:memory recovery contrast}
\ell_{g}(v_{i},t)=-\log\frac{\exp({sim(\hat{\vecs}_{i}(t^+),\vecs_{i}(t^+))})}{\sum_{k=1}^{\left|\mathcal{V}(t^+) \right|}\exp({sim(\hat{\vecs}_{i}(t^+),\vecs_{k}(t^+))})},
\end{equation}
where $\vecs_i(t^+)$ is the current memory vector $\vecs_i$ after processing the interaction at $t$, 
and $\hat{\vecs}_{i}(t^+)$ is the generated memory vector $\hat{\vecs}_{i}$.
We propose a novel self-supervised learning task that encourages a model to learn the normal temporal pattern of data.

\smallsection{Batch Processing for Efficient Training}
For computational efficiency, we employ batch processing for training \method, which has been commonly used~\citep{xu2020inductive,tgn_icml_grl2020,tian2023sad}.
That is, instead of processing a single edge at a time, multiple edges are fed into the model simultaneously.
To this end, a stream of temporal edges is divided into batches of a fixed size in chronological order.
Consequently, memory updates take place at the batch level, and for these updates, the model uses only the interaction information that precedes the current batch.
Note that a single node can be engaged in multiple interactions within a single batch, leading to multiple raw messages for the node. 
To address this, \method aggregates the raw messages into one raw message using mean pooling and continues with the remaining steps of memory update.
In order for the temporal contrast in Eq.~\eqref{eq:memory drift contrast} reflects \textbf{A1}, which emphasizes minimizing the difference between memory vectors before and after an update within a ``short'' time span, the batch size should not be excessively large. 
Therefore, it is crucial to establish an appropriate batch size, taking both this and efficiency into consideration.

\smallsection{Final Training Objective}
For each batch $B$ with temporal edges $\mathcal{E}_{B}$, 
the overall \textit{temporal contrast loss} is defined as follows:
\begin{equation}\label{eq:drift_loss_sum}
\calL_{c}=\frac{1}{\left| \mathcal{E}_B\right|}\sum\nolimits_{(v_{i},v_{j},t)\in \mathcal{E}_B}\omega_{cs}\ell_{c}(v_{i},t)+\omega_{cd}\ell_{c}(v_{j},t),
\end{equation}
where $\omega_{cs}$ and $\omega_{cd}$ are hyperparameters for weighting each source node and each destination node, respectively.
Similarly, the overall \textit{memory generation loss} for $B$ is defined as follows:
\begin{equation}\label{eq:reconstruction_loss_sum}
\calL_{g}=\frac{1}{\left| \mathcal{E}_B\right|}\sum\nolimits_{(v_{i},v_{j},t)\in \mathcal{E}_B}\omega_{gs}\ell_{g}(v_{i},t)+\omega_{gd}\ell_{g}(v_{j},t),
\end{equation}
where $\omega_{gs}$ and $\omega_{gd}$ hyperparameters for weighting each source node and each destination node, respectively.

The final loss $\calL$ for $B$ is the sum of both losses, i.e.,
$\calL = \calL_{c}+\calL_{g}$.
Note that, since anomaly labels are assumed to be unavailable, \method is trained with the assumption that the state of all nodes appearing in the training set is normal, irrespective of their actual states.
However, \method still can learn normal patterns effectively, given that they constitute the majority of the training data, as discussed for various types of anomalies in Section~\ref{sec:discussion}.

\subsection{Anomaly Scoring}\label{sec:method:anomalyscore}
After being trained, \method is able to measure the anomaly score of any node at any given time point. 
\method measures how much each node deviates from \textbf{A1} and \textbf{A2} by computing the \textbf{temporal contrast} score and the \textbf{memory generation} score, which are based on \textbf{A1} and \textbf{A2}, respectively.
Below, we describe each of them.

\smallsection{Temporal Contrast Score}
This score is designed to detect anomalous nodes that deviate from \textbf{A1}, and to this end, it measures the extent of abrupt changes in the long-term interaction pattern.
Specifically, the \textit{temporal contrast score} $sc_{c}(v_{i},t)$ of a node $v_i$ at time $t$ (spec., before processing any interaction at $t$) is defined as the cosine distance between its current and previous memory vectors: 
\begin{equation}\label{eq:drift_score}
sc_{c}(v_{i},t)=1-sim(\vecs_{i}(t),\vecsp_{i}(t)).\\
\end{equation}

\smallsection{Memory Generation Score}
In order to identify anomalous nodes deviating from \textbf{A2}, this score measures the degree of deviation of short-term interaction patterns from long-term interaction patterns.
Specifically, the \textit{memory generation score} $sc_{g}(v_{i},t)$ of a node $v_i$ at time $t$ is defined as the cosine distance between its current and generated memory vectors:
\begin{equation}\label{eq:reconstruct_score}
sc_{g}(v_{i},t)=1-sim(\hat{\vecs}_{i}(t),\vecs_{i}(t)).
\end{equation}

\smallsection{Final Score}
The final anomaly score is a combination of both scores,
i.e., the \textit{final score} $sc(v_{i},t)$ of a node $v_i$ at time $t$ is defined as
\begin{equation}\label{eq:final_score}
sc(v_{i},t)=(sc_{c}(v_{i},t)+sc_{g}(v_{i},t))/4,
\end{equation}
where it is normalized to fall within the range of $[0, 1]$. Nodes in the normal state will have scores closer to $0$, while those in the abnormal state will have scores closer to $1$.

Note that in our notation, $sc(v_{i},t)$ denotes the anomaly score before observing or processing any interaction at time $t$, if such an interaction exists.
We use this score to measure the potential abnormality of node $v_i$'s current state and also that of its subsequent action (which can occur at time $t$ or later).

        \section{Discussion and Analysis}
\label{sec:dis&anlaysis}
In this section, we discuss how \method deals with various types of anomalies.
Then, we analyze the time complexity of \method.

\subsection{Discussion on Anomaly Types}
\label{sec:discussion}
Below, we discuss how \method can detect anomalies of various types, without any prior information about the types.
In Section~\ref{sec:exp:classification} (RQ4), we empirically confirm its effectiveness for all these types.

\smallsection{{T1) Hijacked Anomalies}}
This type involves a previously normal user's account being compromised at some point, exhibiting malicious behaviors that deviate from the user's normal pattern.
\method, motivated by such anomalies, contrasts short-term and long-term interaction patterns to spot them, as discussed above.

\smallsection{T2) New or Rarely-interacting Anomalies}
Many deep learning-based detection models struggle with anomalies involving (1) newly introduced or (2) rarely interacting nodes due to limited data for learning their normal behaviors.
\method assigns a higher anomaly score (both contrast and generation scores) to such nodes because their memory vectors undergo substantial changes until their long-term patterns are established.
We find it advantageous to pay attention to such nodes because, in some of the real-world datasets we used, including Wikipedia, Bitcoin-alpha, and Bitcoin-OTC, new and rarely interacting nodes are more likely to engage in anomalous actions than those with consistent interactions.\footnote{
We suspect that new or inactive accounts are typically used to perform anomalous actions since the cost of being suspended is low for such accounts.}

\smallsection{T3) Consistent Anomalies}
\label{disc:S1}
\method can also effectively identify anomalies exhibiting interaction patterns
that are consistent over time but deviating from those of normal nodes.
Due to the limited capacity (i.e., expressiveness) of the neural networks used in it, \method prioritizes learning prevalent patterns (i.e., those of normal nodes) over less common abnormal ones.
As a result, this can lead to a violation of \textbf{A2}, causing \method to assign high anomaly scores, especially memory generation scores, to such nodes.

\subsection{Complexity Analysis}
\label{sec:analysis}
We analyze the time complexity of \method ``in action'' after being trained. Specifically, we examine the cost of (a) updating the memory in response to a newly arrived edge, and (b) calculating anomaly scores for a query node.

\smallsection{Memory Update}
Given a newly arriving edge, \method updates the memory vector of each endpoint using GRU. 
The total time complexity is dominated by that  of GRU, which is $\calO({d_{s}}^{2}+d_{s}d_{m})$~\citep{rotman2021shuffling}, where $d_{s}$ and $d_{m}$ indicate the dimensions of memory vectors and messages respectively.

\smallsection{Anomaly Scoring}
Given a query node, \method first aims to generate its memory vector using TGAT with input being a zero-masked memory of the query node and the memory vectors of its at most $k$ most recent neighbors.
Thus, the time complexity of obtaining generated memory is $\calO(k{d_{s}}^{2})$~\citep{zheng2023temporal} if we assume, for simplicity, that the embedding dimension of the attention layers is the same as the dimension $d_{s}$ of the memory vectors.
Then, \method computes the similarities (a) between the current memory and previous memory vectors and (b) between current and generated memory vectors, taking $\mathcal{O}({d_{s}}^{2})$ time.
As a result, the overall time complexity of \method for anomaly scoring of a node is $\mathcal{O}(k{d_{s}}^{2})$.

The time complexity for both tasks is $\mathcal{O}(k{d_{s}}^{2}+d_{s}d_{m})$, which is \textbf{constant with respect to the graph size} (i.e., the numbers of accumulated nodes and edges).
If we assume that anomaly scoring (for each endpoint) is performed whenever each edge arrives,
the total complexity becomes linear in the number of accumulated edges, as confirmed empirically in Section~\ref{sec:exp:classification} (RQ2).

\section{Experiments}
\label{sec:exp}

In this section, we review our experiments for answering the following research questions:
\begin{itemize}[leftmargin=*]
\item \textbf{RQ1) Accuracy:} How accurately does \method detect anomalies, compared to state-of-the-art competitors?
\item \textbf{RQ2) Speed:} Does \method exhibit detection speed constant with respect to the graph size?
\item \textbf{RQ3) Ablation Study:} Does every loss and score component of \method contribute to its performance?~\label{RQ:abl}
\item \textbf{RQ4) Type Analysis:} 
Can \method accurately detect various types of anomalies discussed in Section~\ref{sec:discussion}?
\item \textbf {RQ5) Structural Variants:}
How effective is SLADE's model architecture compared to other alternatives?
\end{itemize}

\subsection{Experiment Details}
\label{sec:exp:details}
In this subsection, we describe datasets, baseline methods, and evaluation metrics that are used throughout our experiments. Then, we clarify the implementation details of the proposed method \method.

\smallsection{Datasets}
We assess the performance of \method on four real-world datasets: two social networks (Wikipedia and Reddit~\citep{kumar2019predicting}) and two online financial networks (Bitcoin-alpha and Bitcoin-OTC~\citep{kumar2016edge}).
The Wikipedia dataset records edits made by users on Wikipedia pages.
In this context, when a user is banned after a specific edit, the user's dynamic label is marked as abnormal.
The Reddit dataset consists of posts made by users on subreddits. The user's dynamic label indicates whether the user is banned after the specific post.
The Bitcoin-alpha and Bitcoin-OTC datasets are fundamentally structured as trust-weighted signed networks. Within these datasets, Bitcoin alpha or OTC members assign ratings to other members, ranging from -10 (total distrust) to +10 (complete trust). 
We utilize these ratings with temporal information to identify anomalous nodes and assign dynamic labels to users.
Further details of the dataset processing and the statistics are provided in Appendix~\ref{sec:app:dataset}. Across all datasets, we assess the performance of our method and the baseline models by utilizing the final 15\% of the dataset in chronological order as the test set.

\smallsection{Baselines Methods and Evaluation Metric.}
We extensively compare \method with several baseline methods capable of anomaly detection in CTDGs under inductive settings (i.e., without further optimization for test sets).
For four rule-based methods (SedanSpot~\citep{eswaran2018sedanspot}, MIDAS-R~\citep{bhatia2020midas}, F-FADE~\citep{chang2021f}, Anoedge-l~\citep{bhatia2021sketch}), since they do not require any representation learning process, we evaluate their performance in test sets without model training.
For five neural network-based models (JODIE~\citep{kumar2019predicting}, Dyrep~\citep{trivedi2019dyrep}, TGAT~\citep{xu2020inductive}, TGN~\citep{tgn_icml_grl2020}, SAD~\citep{tian2023sad}), we train each model by using train sets and do hyperparameter tuning with validation sets. 
At last, we evaluate them by using test sets.
At this point, excluding SAD, neural network-based models follow the previous approach of training the encoder through link prediction-based self-supervised learning and subsequently training the classifier with labels through supervised learning.
In scenarios where a model requires node or edge features, but these attributes are absent from the dataset, the model uses zero vectors as substitutes.
A detailed description of these baselines is provided in Online Appendix C.
To quantify the performance of each model, we use Area Under ROC (AUC) as an evaluation metric. For extra results in terms of Average Precision (AP), refer to Online Appendix D.1.

\smallsection{Implementation Details}
In our experiments, we use two versions of our method, \method and \method-HP. 
For \method, as label utilization is not possible in its original setting, we do not conduct hyperparameter tuning on the validation dataset split. 
Instead, we use the same hyperparameter combination across all datasets.
We adopt a batch size of $100$ and an initial learning rate of $3\times10^{-6}$. we also fix a memory dimension to $256$, a message dimension to $128$, and a max number of most recent neighbors (node degree) to $20$. 
For the loss function, we set $\omega_{cs}=1, \omega_{cd}=1, \omega_{gs}=0.1$, and $\omega_{gd}=0.1$ across all datasets.
We train the model on the training set for 10 epochs and then evaluate its performance on the test set.

Another version of \method is \textbf{\method-HP}, where we tune the hyperparameters of \method by using the validation split of each dataset, just as we tune the hyperparameters of ``all'' baseline methods, including the rule-based ones.
Refer to Appendix~\ref{sec:app:impl:model_param} for details regarding the hyperparameter search space of \method-HP. 

For the baseline methods, we assess the performance on the test set using two hyperparameter settings:
(a) those recommended by the authors, and (b) those leading to the best validation performance for each specific dataset. 
Subsequently, we conduct a performance evaluation on the test set using these two distinct settings and report the best-performing outcome.

The reported results have been obtained by averaging across 10 separate runs, each with different random initializations of the models.
A further description of the implementation and hyperparameters of \method and baselines are provided in Appendix~\ref{sec:app:impl}.

\subsection{Experimental Results}
\label{sec:exp:classification}

\begin{table}[!t]
    \centering
    \caption{\label{tab:AUC} AUC (in \%) in the detection of dynamic anomaly nodes. The first four methods are rule-based models, and the others are based on representation learning. 
    For each dataset, the best and the second-best performances are highlighted in \textbf{boldface} and \ul{underlined}, respectively.
    In most cases, \method and \method-HP perform best, even when compared to models that rely on label information.
    For results in terms of Average Precision (AP), refer to Online Appendix D.1.
    }
    \setlength{\tabcolsep}{2.5pt}
    \small
    \scalebox{0.9}{
        \def\arraystretch{1.0}
        \begin{tabular}{l|cccc}
            \toprule
            Method & Wikipedia & Reddit & Bitcoin-alpha & Bitcoin-OTC  \\
            \midrule
            \midrule
                \SedanSpot~{\citep{eswaran2018sedanspot}}  & 82.88 $\pm$ 1.54   & 58.97 $\pm$ 1.85  & 69.09 $\pm$ 0.76        & 71.71 $\pm$ 0.73     \\
                \MIDAS~{\citep{bhatia2020midas}}  & 62.92 $\pm$ 3.53   & 59.94 $\pm$ 0.95  & 64.57 $\pm$ 0.11        & 62.16 $\pm$ 1.99    \\
                \FFADE~{\citep{chang2021f}}  & 44.88 $\pm$ 0.00   & 49.79 $\pm$ 0.05  & 53.57 $\pm$ 0.00        & 51.12 $\pm$ 0.00      \\
                \Anoedgel~{\citep{bhatia2021sketch}}            & 47.38 $\pm$ 0.32   & 48.47 $\pm$ 0.46  & 62.52 $\pm$ 0.24        & 65.99 $\pm$ 0.19    \\
            \midrule 
                \JODIE~{\citep{kumar2019predicting}} & 85.75 $\pm$ 0.17   & 61.47 $\pm$ 0.58   & 73.53 $\pm$ 1.26  & 69.13 $\pm$ 0.96   \\
                \Dyrep~{\citep{trivedi2019dyrep}} & 85.68 $\pm$ 0.34   & 63.42 $\pm$ 0.62   & 73.30 $\pm$ 1.51   & 70.76 $\pm$ 0.70   \\
                \TGAT~{\citep{xu2020inductive}}  & 83.24 $\pm$ 1.11   & 65.12 $\pm$ 1.65   & 71.67 $\pm$ 0.90   & 68.33 $\pm$ 1.23   \\
                \TGN~{\citep{tgn_icml_grl2020}}  & 87.47 $\pm$ 0.22   & 67.16 $\pm$ 1.03   & 69.90 $\pm$ 0.99   & \ul{76.23 $\pm$ 0.23}     \\
                \SAD~{\citep{tian2023sad}}   & 86.15 $\pm$ 0.63       & 68.45 $\pm$ 1.27       & 68.56 $\pm$ 3.17       & 64.67 $\pm$ 4.97 \\
            \midrule
            \rule{0pt}{8pt}
                \tb{\method}
                 &\ul{87.75 $\pm$ 0.68}  & \ul{72.19 $\pm$ 0.60}  & \ul{76.32 $\pm$ 0.28}  & 75.80 $\pm$ 0.19   \\
                 \rule{0pt}{10pt}
                \tb{\method-HP}        & \tb{88.68 $\pm$ 0.39}  & \tb{75.08 $\pm$ 0.50}  & \tb{76.92 $\pm$ 0.36}  & \tb{77.18 $\pm$ 0.27} \rule{0pt}{10pt}   \\
            \bottomrule
        \end{tabular}
}
    \vspace{-0.0mm}
\end{table}



\begin{figure}[!t]
    \centering
    \includegraphics[width=0.48\textwidth]
    {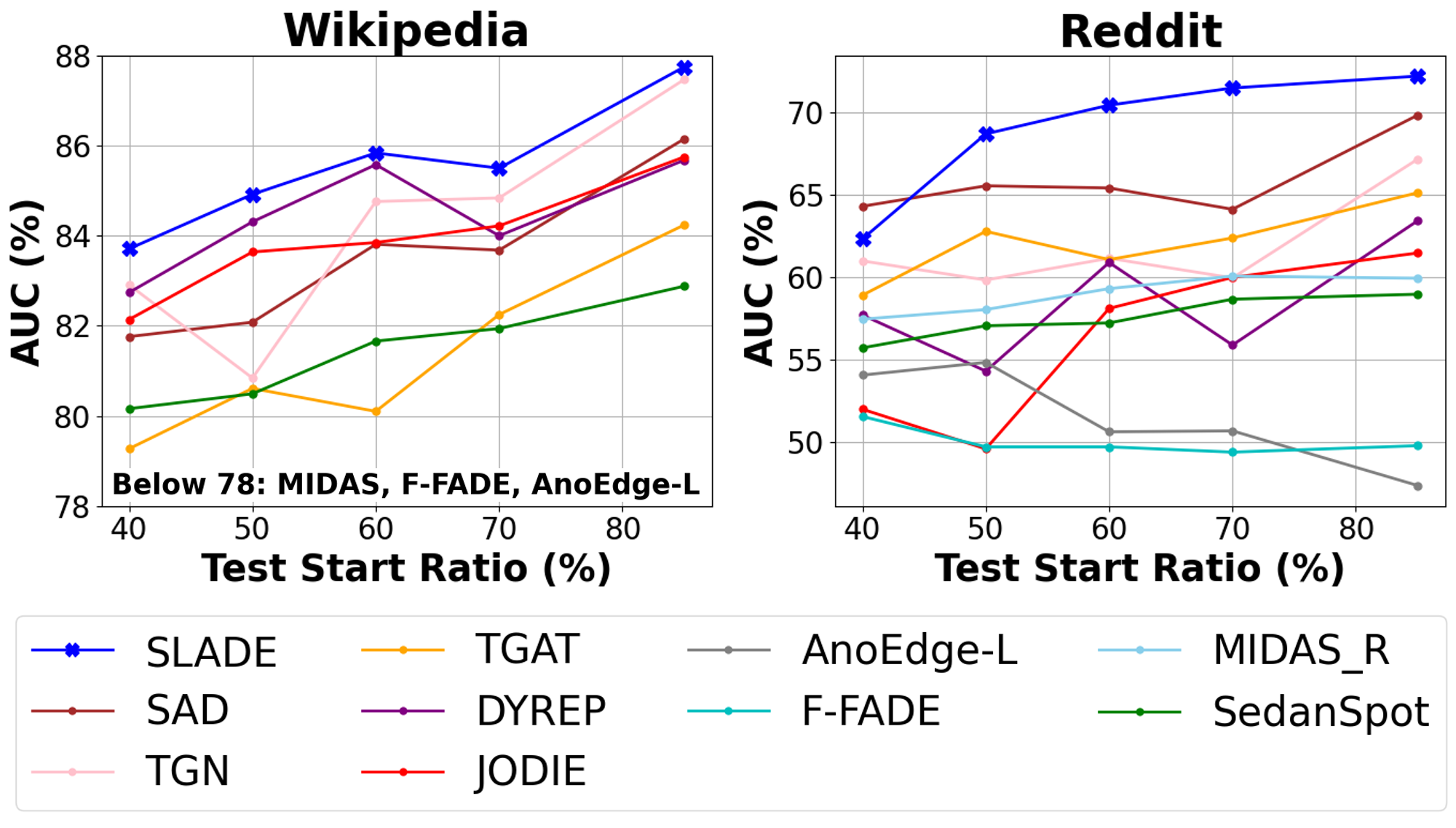}
    \caption{ \label{fig:test_start_ratio}
        AUC (in \%) when varying the test start ratio. For learning-based methods, temporal edges preceding the test start ratio in the dataset are employed for training. If validation is needed, the last 10\% of the training set is used for validation.
        Note that \method performs best in most cases.
    }
\end{figure}

\smallsection{RQ1) Accuracy}
As shown in Table~\ref{tab:AUC}, \method and \method-HP significantly outperform other baseline methods in most of the datasets.
There are two notable observations in this analysis.

First, regarding the unsupervised learning competitors (i.e., \SedanSpot, \MIDAS, \FFADE, and \Anoedgel), \method consistently outperforms them across all datasets, achieving performance gains of up to 20.44\% (in the Reddit dataset) compared to the second-best performing unsupervised model.
This result demonstrates that real-world anomalies exhibit complex patterns hard to be fully captured by fixed rule-based approaches.
As a result, the necessity for more intricate representation models becomes evident.

Second, interestingly, while \method does not utilize any label information, even for hyperparameter tuning, it still outperforms all supervised baseline models, on all datasets except for the BitcoinOTC dataset.
This result implies that the patterns of nodes in the normal state in real-world graphs closely adhere to
\textbf{A1} and \textbf{A2}, and \method effectively captures such patterns.
Furthermore, it is evident that measuring the extent to which nodes deviate from the patterns provides crucial information for anomaly detection.

In addition, we measure the performances of the considered methods while varying the proportion of the training split (or equivalently the test split).
Specifically, we utilize the first $\calT\%$ of the edges as a train set and assess each model by using the remaining $(100 - \calT)\%$ of edges.
We refer to $\calT$ as a \textit{test start ratio}.
As shown in Figure~\ref{fig:test_start_ratio}, \method outperforms all baseline methods in most of the settings, across various test start ratios.
Moreover, as seen in the performances of \method between the 40\% and 80\% ratios in the Wikipedia dataset, using just half of the dataset leads to only a marginal performance degradation (spec., 3.2\%).

\begin{figure}[!t]
    \centering
    \setlength\aboverulesep{0.5pt}
    \setlength\belowrulesep{0.5pt}
    \centering
    \includegraphics[width=1\linewidth]{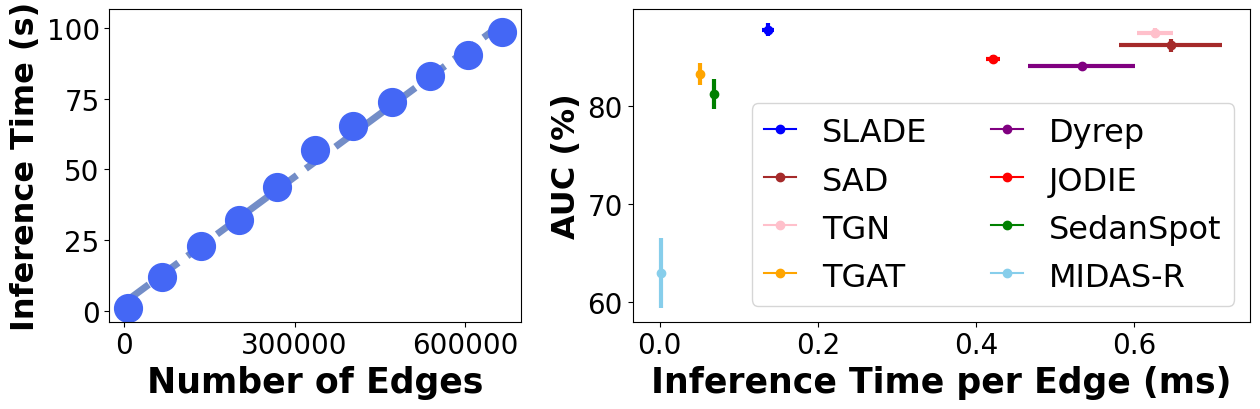} 
    \caption{
    \label{fig:inference_time}
         The left figure shows the linear increase of the running time of \method with respect to the number of edges in the Reddit dataset. The right figure shows the trade-off between detection speed and accuracy (with standard deviations) in the Wikipedia dataset provided by the competing methods.
         The baseline methods with AUC scores below 60\% are excluded from consideration to enhance the clarity of performance differences between the methods.
         \method exhibits constant processing time per edge  (as proven in Section~\ref{sec:analysis}), offering the best trade-off between speed and accuracy. For a training-time comparison, refer to Online Appendix D.2.
    }
\end{figure}

\smallsection{RQ2) Speed in Action}
\label{sec:exp:clustering}
To empirically demonstrate the theoretical complexity analysis in Section~\ref{sec:analysis}, 
we measure the running time of \method ``in action'' after being trained on the Reddit dataset while varying the number of edges.
Anomaly scoring is performed (for each endpoint) only when each edge arrives.
As depicted in the left subplot of Figure~\ref{fig:inference_time}, the running time of \method is linear in the number of edges, being aligned with our analysis.

Additionally, we compare the running time and AUC scores of \method with those of all considered methods.
As shown in the right plot of Figure~\ref{fig:inference_time}, while \method is about 1.99$\times$ slower than SedanSpot, which is the most accurate rule-based approach, \method demonstrates a significant improvement of 5.88\% in AUC scores, compared to SedanSpot.
Furthermore, \method is about 4.57$\times$ faster than TGN, which is the second most accurate following \method.

\begin{table}
    \setlength{\tabcolsep}{5pt}
    \setlength\aboverulesep{2pt}
    \centering
    \caption{\label{tab:comparison}Comparison of AUC (in \%) of \method and its variants that use a subset of the proposed components (i.e., temporal contrast loss $\calL_{c}$, memory generation loss $\calL_{g}$, temporal contrast score $sc_{c}$, and memory generation score $sc_{g}$).
    \method with all components performs the best overall, showing the effectiveness of each.}
    \small
    \scalebox{0.8}{
        \def\arraystretch{1.0}
        \begin{tabular}{cccccccc}
            \toprule
            $\calL_{c}$ & $\calL_{g}$ & $sc_{c}$ & $sc_{g}$ & Wikipedia & Reddit & Bitcoin-alpha & Bitcoin-OTC \\
            \midrule
            \midrule
            \ding{51}   & -    & \ding{51}      & -         & 85.69 $\pm$ 1.17       & 48.89 $\pm$ 1.53      & \ul{76.38 $\pm$ 0.48}       & 73.87 $\pm$ 0.23       \\
            -           & \ding{51} & -     & \ding{51}    & \ul{87.57 $\pm$ 0.35}       & 61.77 $\pm$ 0.12      & 74.84 $\pm$ 0.40       & \bf{75.80 $\pm$ 0.26}          \\
            \ding{51}           & \ding{51}         & \ding{51} & - & 87.42 $\pm$ 1.06       & 48.51 $\pm$ 1.56       & \bf{76.45 $\pm$ 0.52}       & 74.10 $\pm$ 0.25         \\
            \ding{51}   & \ding{51} & -   & \ding{51}      & 84.64 $\pm$ 0.75  & \ul{71.33 $\pm$ 0.56}       & 75.24 $\pm$ 3.50       & 75.70 $\pm$ 0.21   \\
            \midrule
            \ding{51}   & \ding{51}  & \ding{51} & \ding{51} & \bf{87.75 $\pm$ 0.68}       & \bf{72.19 $\pm$ 0.60}       & 76.32 $\pm$ 0.28  & \bf{75.80 $\pm$ 0.19}       \\
            \bottomrule
        \end{tabular}
    }
    \vspace{-1.5mm}
\end{table}
\smallsection{RQ3) Ablation Study}
\label{sec:exp:tsne}
The ablation study is conducted across the four datasets to analyze the necessity of the used self-supervised losses (Eq~\eqref{eq:memory drift contrast} and Eq~\eqref{eq:memory recovery contrast}) and two anomaly detection scores (Eq~\eqref{eq:drift_score} and Eq~\eqref{eq:reconstruct_score}).
To this end, we utilize several variants of \method, where certain scores or self-supervised losses are removed from \method.

As evident from Table~\ref{tab:comparison}, \method, which uses all the proposed self-supervised losses and scores, outperforms its variants in most of the datasets  (Wikipedia, Reddit, and Bitcoin-OTC).
Moreover, even in the Bitcoin-alpha dataset, the performance gap between \method and the best-performing variant is within the standard deviation range.
Notably, the generation score greatly contributes to accurate anomaly detection in most of the dataset, providing an empirical performance gain of up to 48.81\% (on the Reddit dataset).

\begin{table}[!t]
    \vspace{-2mm}
    \centering
    \caption{\label{tab:Email_AUC} AUC (in \%) in the detection of dynamic anomaly nodes in the two synthetic datasets. 
    Since anomalies are injected only into the test sets, the comparison is limited to unsupervised methods. \FFADE, which consistently achieves an AUC value below 0.5, is omitted from the table. For results of Average Precision (AP), refer to Online Appendix D.1.
    }
    \small
    \scalebox{0.85}{
        \def\arraystretch{1.2}
        \begin{tabular}{l|ccc|c}
            \toprule
            Dataset & \SedanSpot~\citep{eswaran2018sedanspot} & \MIDAS~\citep{bhatia2020midas} &  \Anoedgel~\citep{bhatia2021sketch} & \method  \\
            \midrule
            \midrule                
            Synthetic-Hijack & 77.13 $\pm$ 2.21   & \ul{81.80 $\pm$ 1.26} &  58.87 $\pm$ 2.47 & \textbf{98.08 $\pm$ 1.02}     \\ 
            Synthetic-New  & 78.05 $\pm$ 1.68   & \ul{82.63 $\pm$ 0.07} &  61.86 $\pm$ 2.41 & \textbf{98.38 $\pm$ 1.09}    \\           
            \bottomrule
        \end{tabular}
}
\end{table}

\smallsection{RQ4) Type Analysis}
\label{sec:exp:email}
We demonstrate the effectiveness of \method in capturing various types of anomalies discussed in Section~\ref{sec:discussion}. To this end, we create synthetic datasets by injecting anomalies to the Email-EU~\citep{paranjape2017motifs} dataset, which consists of emails between users in a large European research institution.
Specifically, we create anomalies that repetitively send spam emails to random recipients within a short time interval, mimicking spammers, and based on the timing of spamming, we have two different datasets: 
\begin{itemize}[leftmargin=*]
\item \textbf{Synthetic-Hijack}: Accounts of previously normal users start disseminating spam emails after being hijacked at a certain time point and continue their anomalous actions.
We further categorize each anomaly as \textbf{Type T1} during its initial 20 interactions after being hijacked, and as \textbf{Type T3} after that.
\item \textbf{Synthetic-New}: New anomalous users appear and initiate spreading spam emails from the beginning and continue their anomalous actions.
We further categorize each anomaly as \textbf{Type T2} during its first 20 interactions, and as \textbf{Type T3} afterward.
\end{itemize}
In them, all anomalies are injected only into the test set (i.e., final $10\%$ of the dataset in chronological order), ensuring that they remain unknown to the model during training.
Consequently, the task involves detecting these anomalies in an unsupervised manner, and  \method is compared only with the unsupervised baseline methods.

\begin{figure}[!t]
    \centering
    \setlength\aboverulesep{0.5pt}
    \setlength\belowrulesep{0.5pt}
    \centering
    \includegraphics[width=1\linewidth]{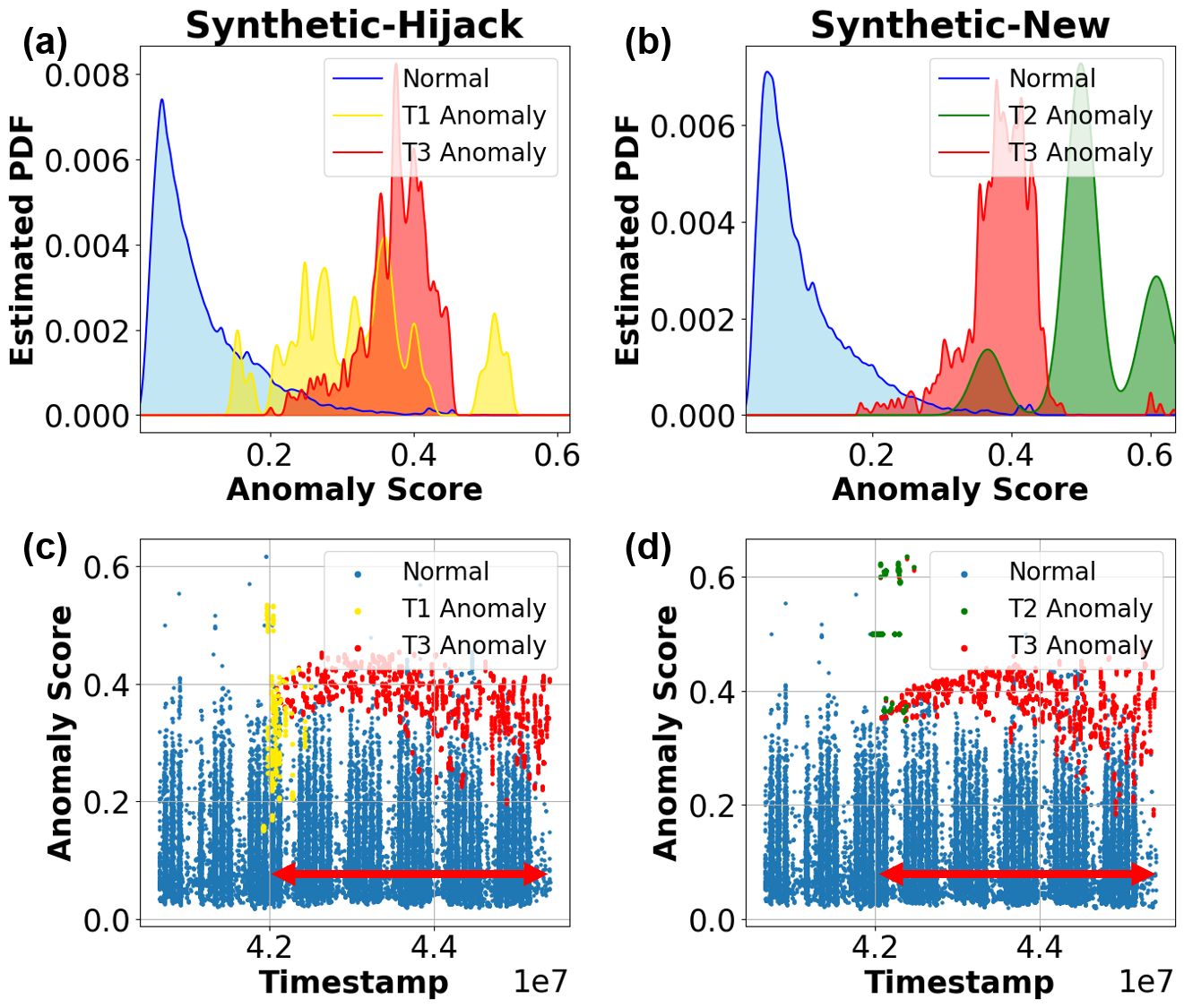} 
    \caption{ \label{fig:Email_test}
         (a) and (b) show the distribution of anomaly scores assigned by \method to instances of each node type in the two synthetic datasets (visualization is based on Gaussian kernel density estimation).
         (c) and (d) show the anomaly scores at each time period.
         Note that in all figures, \method clearly distinguishes anomalies from normal nodes.
         For results from several baseline methods, refer to Online Appendix D.3. 
    }
\end{figure}

As shown in Table~\ref{tab:Email_AUC}, \method performs best, achieving performance gains up to 19.9\% in the Synthetic-Hijack dataset and 19.06\% in the Synthetic-New dataset.
These results reaffirm the limitations of traditional unsupervised methods in capturing anomalies beyond their targeted types.
In contrast, as it can learn normal patterns from data, \method successfully identifies various types of anomalies.

Furthermore, we provide qualitative analysis of how \method assigns scores to the normal and anomalies of each type.
Figures~\ref{fig:Email_test}(a) and (b) show that \method clearly separates anomaly score distributions of all anomaly types (\textbf{T1}, \textbf{T2}, and \textbf{T3}) from the distribution of the normal ones.
Figure~\ref{fig:Email_test}(c) shows that
the anomaly scores of hijacked anomalies (\textbf{T1}) increase shortly after being hijacked.
Figure~\ref{fig:Email_test}(d) shows that \method successfully assigns high anomaly scores to new or rarely interacting anomalies (\textbf{T2}).
Consistent anomalies (\textbf{T3}) receive high anomaly scores in both cases.

\smallsection{RQ5) Structural Variants}
\label{sec:app:exp:strucutre}
There are two major neural network components in \method: (1) \textbf{GRU}~\citep{chung2014empirical}, which updates the memory of each node, and (2) \textbf{TGAT}~\citep{xu2020inductive}, which generates the memory of a target node.
To demonstrate the effectiveness of each module in dynamic anomaly detection in edge stream, we compare the performances of \method and its several variants where the memory updater and the memory generator are replaced by other neural network architectures.

\begin{itemize}[leftmargin=*]
    \item \textbf{\method-MLP (Instead of GRU)}: In this variant, we use an MLP instead of the GRU module to update the memory of each node.
    Specifically, the memory update procedure (Eq (3) in the main paper) is replaced by $\vecs_{i} = \text{MLP}([\vecm_{i} \vert \vert \vecs_{i}]).$
    Note that in this variant, $\vecm_{i}$ and $\vecs_{i}$ are treated as if they are independent, and thus it cannot capture the temporal dependency between them.     
    \item \textbf{\method-GAT (Instead of TGAT)}: This variant calculates attention scores based only on the currently given memory information, as follows:
\begin{align}\label{eq:GAT}
& \hat{\vecs}_{i}=\text{MultiHeadAttention}(\vecq,\textbf{K},\textbf{V}), \ \vecq=\vecs_{i},\\
& \textbf{K}=\textbf{V} = \left [ \vecs_{n_1},...,\vecs_{n_k} \right],  \nonumber
\end{align}
    where $\{n_{1}, ..., n_{k}\}$ denote the indices of the neighbors of the target node $v_i$.
    Note that this variant cannot incorporate temporal information in its attention mechanism.
    \item \textbf{\method-SUM (Instead of TGAT)}: In this variant, we use a modified temporal graph sum~\citep{tgn_icml_grl2020} for message passing, as follows:
\begin{align}\label{eq:sum}
& \hat{\vecs}_{i}=\textbf{W}_{2}(\left [\bar{\vecs}_{i}||\phi(t-t)  \right ]), \\
& \bar{\vecs}_{i}=\text{ReLU}(\sum_{j=1}^{k}\textbf{W}_{1}(\vecs_{n_j}||\phi(t-t'_{n_j}))), 
\nonumber
\end{align}
    where $\textbf{W}_{1},\textbf{W}_{2}\in \mathbb{R}^{d_s\times 2d_s}$, $\{n_{1}, ..., n_{k}\}$ denote the indices of the neighbors of the target node $v_i$, $\{t'_{n_1}, ..., t'_{n_k}\}$ denote the times of the most recent interactions with them, and the weights of each linear layer are denoted as $\textbf{W}_{1}$ and $\textbf{W}_{2}$, respectively.
    Note that this variant does not use any attention mechanism i.e., all neighbors are treated with equal importance.
\end{itemize}

\begin{table}[!t]
    \centering
    \caption{Comparison of AUC (in \%) of SLADE and its structural variants. For each dataset, the best and the second-best performances are highlighted in \textbf{boldface} and \ul{underlined}, respectively.
    Across all datasets, \method consistently demonstrated the best or second-best performance compared to the other variants.
    }
    \setlength{\tabcolsep}{2.5pt}
    \small
    \scalebox{0.9}{
        \def\arraystretch{1.0}
        \begin{tabular}{l|cccc}
            \toprule
            Method & Wikipedia & Reddit & Bitcoin-alpha & Bitcoin-OTC  \\
            \midrule
            \midrule
                \method-MLP   & 85.97 $\pm$ 1.17   & 55.35 $\pm$ 3.34  & 74.68 $\pm$ 1.35        & \ul{75.42 $\pm$ 0.65}     \\
            \midrule 
                \method-GAT & 86.86 $\pm$ 0.44   & 67.15 $\pm$ 1.85   & 75.67 $\pm$ 0.57  & 74.24 $\pm$ 0.25   \\
                \method-SUM & \textbf{88.43 $\pm$ 0.44}   & \ul{70.89 $\pm$ 0.53}   & \ul{75.84 $\pm$ 0.26}   & \ul{75.42 $\pm$ 0.15}   \\
            \midrule
            \rule{0pt}{8pt}
                \tb{\method}
                 &\ul{87.75 $\pm$ 0.68}  & \textbf{72.19 $\pm$ 0.60}  & \textbf{76.32 $\pm$ 0.28}  & \textbf{75.80 $\pm$ 0.19}   \\
            \bottomrule
        \end{tabular}
    }
    \normalsize
    \label{tab:variants}
\end{table}

We compare the performances of \method and the above three variants, utilizing the same hyperparameter settings as the original \method.
As shown in Table~\ref{tab:variants}, \method achieves the best performance in three out of four datasets.
This result demonstrates the effectiveness of GRU and TGAT, i.e., the importance of modeling temporal dependency in memory update and temporal attention in memory generation.
Specifically, \method-MLP and \method-GAT consistently underperform \method and \method-SUM, demonstrating the importance of utilizing temporal information and temporal dependency in detecting dynamic anomalies.
While \method-SUM outperforms \method in the Wikipedia dataset, its performance gain is marginal, falling within the standard deviation.

\smallsection{Extra Experimental Results in  Online Appendix~\citep{Lee_SLADE_Online_Appendix_2024}}
We explore more baselines, including (a) neural networks trained through link-prediction-based self-supervised learning and (b) anomaly-detection methods based on DTDGs and static graphs.
We also use a transportation network dataset for evaluation. We further investigate (a) robustness to anomalies in training, (b) training with label supervision, (c) learnable time encoding, (d) anomalies with camouflage, and (e) dynamic heterogeneous graphs.

        \vspace{-2mm}

	\section{Conclusion}
\label{sec:conclusion}

We proposed \method, a novel self-supervised method for dynamic anomaly detection in edge streams, with the following strengths:
\begin{itemize}[leftmargin=10pt,itemsep=1pt,topsep=0pt]
    \item \method does not rely on any label supervision while being able to capture complex anomalies (Section~\ref{sec:method}).
    \item \method outperforms state-of-the-art anomaly detection methods in the task of dynamic anomaly detection in edge streams. \method achieves a performance improvement of up to 12.80\% and 4.23\% compared to the best-performing unsupervised and supervised baseline methods, respectively (Section~\ref{sec:exp:classification}).
    \item \method demonstrates a constant time complexity per edge, which is theoretically supported (Section~\ref{sec:analysis}). 
\end{itemize}

        \subsection*{Acknowledgements}
        {\small This work was supported by Institute of Information \& Communications Technology Planning \& Evaluation (IITP) grant funded by the Korea government (MSIT)  (No. 2022-0-00157, Robust, Fair, Extensible Data-Centric Continual Learning) (No. RS-2019-II190075, Artificial Intelligence Graduate School Program (KAIST)).
        }
        \bibliographystyle{ACM-Reference-Format}
        \balance
	\bibliography{ref}


\begin{thebibliography}{42}


\ifx \showCODEN    \undefined \def \showCODEN     #1{\unskip}     \fi
\ifx \showDOI      \undefined \def \showDOI       #1{#1}\fi
\ifx \showISBNx    \undefined \def \showISBNx     #1{\unskip}     \fi
\ifx \showISBNxiii \undefined \def \showISBNxiii  #1{\unskip}     \fi
\ifx \showISSN     \undefined \def \showISSN      #1{\unskip}     \fi
\ifx \showLCCN     \undefined \def \showLCCN      #1{\unskip}     \fi
\ifx \shownote     \undefined \def \shownote      #1{#1}          \fi
\ifx \showarticletitle \undefined \def \showarticletitle #1{#1}   \fi
\ifx \showURL      \undefined \def \showURL       {\relax}        \fi
\providecommand\bibfield[2]{#2}
\providecommand\bibinfo[2]{#2}
\providecommand\natexlab[1]{#1}
\providecommand\showeprint[2][]{arXiv:#2}

\bibitem[Aalen et~al\mbox{.}(2008)]%
        {aalen2008survival}
\bibfield{author}{\bibinfo{person}{Odd Aalen}, \bibinfo{person}{Ornulf Borgan}, {and} \bibinfo{person}{Hakon Gjessing}.} \bibinfo{year}{2008}\natexlab{}.
\newblock \bibinfo{booktitle}{\emph{Survival and event history analysis: a process point of view}}.
\newblock \bibinfo{publisher}{Springer Science \& Business Media}.
\newblock


\bibitem[Anderson et~al\mbox{.}(2014)]%
        {anderson2014dynamics}
\bibfield{author}{\bibinfo{person}{Ashton Anderson}, \bibinfo{person}{Ravi Kumar}, \bibinfo{person}{Andrew Tomkins}, {and} \bibinfo{person}{Sergei Vassilvitskii}.} \bibinfo{year}{2014}\natexlab{}.
\newblock \showarticletitle{The dynamics of repeat consumption}. In \bibinfo{booktitle}{\emph{WWW}}.
\newblock


\bibitem[Benson et~al\mbox{.}(2018)]%
        {benson2018sequences}
\bibfield{author}{\bibinfo{person}{Austin~R Benson}, \bibinfo{person}{Ravi Kumar}, {and} \bibinfo{person}{Andrew Tomkins}.} \bibinfo{year}{2018}\natexlab{}.
\newblock \showarticletitle{Sequences of sets}. In \bibinfo{booktitle}{\emph{KDD}}.
\newblock


\bibitem[Bhatia et~al\mbox{.}(2020)]%
        {bhatia2020midas}
\bibfield{author}{\bibinfo{person}{Siddharth Bhatia}, \bibinfo{person}{Bryan Hooi}, \bibinfo{person}{Minji Yoon}, \bibinfo{person}{Kijung Shin}, {and} \bibinfo{person}{Christos Faloutsos}.} \bibinfo{year}{2020}\natexlab{}.
\newblock \showarticletitle{Midas: Microcluster-based detector of anomalies in edge streams}. In \bibinfo{booktitle}{\emph{AAAI}}.
\newblock


\bibitem[Bhatia et~al\mbox{.}(2023)]%
        {bhatia2021sketch}
\bibfield{author}{\bibinfo{person}{Siddharth Bhatia}, \bibinfo{person}{Mohit Wadhwa}, \bibinfo{person}{Kenji Kawaguchi}, \bibinfo{person}{Neil Shah}, \bibinfo{person}{Philip~S Yu}, {and} \bibinfo{person}{Bryan Hooi}.} \bibinfo{year}{2023}\natexlab{}.
\newblock \showarticletitle{Sketch-Based Anomaly Detection in Streaming Graphs}. In \bibinfo{booktitle}{\emph{KDD}}.
\newblock


\bibitem[Chang et~al\mbox{.}(2021)]%
        {chang2021f}
\bibfield{author}{\bibinfo{person}{Yen-Yu Chang}, \bibinfo{person}{Pan Li}, \bibinfo{person}{Rok Sosic}, \bibinfo{person}{MH Afifi}, \bibinfo{person}{Marco Schweighauser}, {and} \bibinfo{person}{Jure Leskovec}.} \bibinfo{year}{2021}\natexlab{}.
\newblock \showarticletitle{F-fade: Frequency factorization for anomaly detection in edge streams}. In \bibinfo{booktitle}{\emph{WSDM}}.
\newblock


\bibitem[Cho et~al\mbox{.}(2014)]%
        {cho2014learning}
\bibfield{author}{\bibinfo{person}{Kyunghyun Cho}, \bibinfo{person}{Bart Van~Merri{\"e}nboer}, \bibinfo{person}{Caglar Gulcehre}, \bibinfo{person}{Dzmitry Bahdanau}, \bibinfo{person}{Fethi Bougares}, \bibinfo{person}{Holger Schwenk}, {and} \bibinfo{person}{Yoshua Bengio}.} \bibinfo{year}{2014}\natexlab{}.
\newblock \showarticletitle{Learning phrase representations using RNN encoder-decoder for statistical machine translation}. In \bibinfo{booktitle}{\emph{EMNLP}}.
\newblock


\bibitem[Chung et~al\mbox{.}(2014)]%
        {chung2014empirical}
\bibfield{author}{\bibinfo{person}{Junyoung Chung}, \bibinfo{person}{Caglar Gulcehre}, \bibinfo{person}{KyungHyun Cho}, {and} \bibinfo{person}{Yoshua Bengio}.} \bibinfo{year}{2014}\natexlab{}.
\newblock \showarticletitle{Empirical evaluation of gated recurrent neural networks on sequence modeling}. In \bibinfo{booktitle}{\emph{NeurIPS, Deep Learning and Representation Learning Workshop}}.
\newblock


\bibitem[Cong et~al\mbox{.}(2022)]%
        {cong2022we}
\bibfield{author}{\bibinfo{person}{Weilin Cong}, \bibinfo{person}{Si Zhang}, \bibinfo{person}{Jian Kang}, \bibinfo{person}{Baichuan Yuan}, \bibinfo{person}{Hao Wu}, \bibinfo{person}{Xin Zhou}, \bibinfo{person}{Hanghang Tong}, {and} \bibinfo{person}{Mehrdad Mahdavi}.} \bibinfo{year}{2022}\natexlab{}.
\newblock \showarticletitle{Do We Really Need Complicated Model Architectures For Temporal Networks?}. In \bibinfo{booktitle}{\emph{ICLR}}.
\newblock


\bibitem[Ding et~al\mbox{.}(2019)]%
        {ding2019deep}
\bibfield{author}{\bibinfo{person}{Kaize Ding}, \bibinfo{person}{Jundong Li}, \bibinfo{person}{Rohit Bhanushali}, {and} \bibinfo{person}{Huan Liu}.} \bibinfo{year}{2019}\natexlab{}.
\newblock \showarticletitle{Deep anomaly detection on attributed networks}. In \bibinfo{booktitle}{\emph{SDM}}.
\newblock


\bibitem[Ding et~al\mbox{.}(2021)]%
        {ding2021few}
\bibfield{author}{\bibinfo{person}{Kaize Ding}, \bibinfo{person}{Qinghai Zhou}, \bibinfo{person}{Hanghang Tong}, {and} \bibinfo{person}{Huan Liu}.} \bibinfo{year}{2021}\natexlab{}.
\newblock \showarticletitle{Few-shot network anomaly detection via cross-network meta-learning}. In \bibinfo{booktitle}{\emph{WWW}}.
\newblock


\bibitem[Eswaran and Faloutsos(2018)]%
        {eswaran2018sedanspot}
\bibfield{author}{\bibinfo{person}{Dhivya Eswaran} {and} \bibinfo{person}{Christos Faloutsos}.} \bibinfo{year}{2018}\natexlab{}.
\newblock \showarticletitle{Sedanspot: Detecting anomalies in edge streams}. In \bibinfo{booktitle}{\emph{ICDM}}.
\newblock


\bibitem[Huang et~al\mbox{.}(2022)]%
        {huang2022dynamic}
\bibfield{author}{\bibinfo{person}{Jin Huang}, \bibinfo{person}{Wentai Zhu}, \bibinfo{person}{Jing Xiao}, \bibinfo{person}{Tian Lu}, {and} \bibinfo{person}{Weihao Yu}.} \bibinfo{year}{2022}\natexlab{}.
\newblock \showarticletitle{Dynamic Graph Representation Based on Temporal and Contextual Contrasting}. In \bibinfo{booktitle}{\emph{ACAI}}.
\newblock


\bibitem[Jin et~al\mbox{.}(2021)]%
        {jin2021anemone}
\bibfield{author}{\bibinfo{person}{Ming Jin}, \bibinfo{person}{Yixin Liu}, \bibinfo{person}{Yu Zheng}, \bibinfo{person}{Lianhua Chi}, \bibinfo{person}{Yuan-Fang Li}, {and} \bibinfo{person}{Shirui Pan}.} \bibinfo{year}{2021}\natexlab{}.
\newblock \showarticletitle{Anemone: Graph anomaly detection with multi-scale contrastive learning}. In \bibinfo{booktitle}{\emph{CIKM}}.
\newblock


\bibitem[Kazemi et~al\mbox{.}(2020)]%
        {kazemi2020representation}
\bibfield{author}{\bibinfo{person}{Seyed~Mehran Kazemi}, \bibinfo{person}{Rishab Goel}, \bibinfo{person}{Kshitij Jain}, \bibinfo{person}{Ivan Kobyzev}, \bibinfo{person}{Akshay Sethi}, \bibinfo{person}{Peter Forsyth}, {and} \bibinfo{person}{Pascal Poupart}.} \bibinfo{year}{2020}\natexlab{}.
\newblock \showarticletitle{Representation learning for dynamic graphs: A survey}.
\newblock \bibinfo{journal}{\emph{Journal of Machine Learning Research}} \bibinfo{volume}{21}, \bibinfo{number}{1} (\bibinfo{year}{2020}), \bibinfo{pages}{2648--2720}.
\newblock


\bibitem[Khoshraftar et~al\mbox{.}(2022)]%
        {khoshraftar2022temporal}
\bibfield{author}{\bibinfo{person}{Shima Khoshraftar}, \bibinfo{person}{Aijun An}, {and} \bibinfo{person}{Nastaran Babanejad}.} \bibinfo{year}{2022}\natexlab{}.
\newblock \showarticletitle{Temporal graph representation learning via maximal cliques}. In \bibinfo{booktitle}{\emph{Big Data}}.
\newblock


\bibitem[Kingma and Ba(2014)]%
        {kingma2014adam}
\bibfield{author}{\bibinfo{person}{Diederik~P Kingma} {and} \bibinfo{person}{Jimmy Ba}.} \bibinfo{year}{2014}\natexlab{}.
\newblock \showarticletitle{Adam: A method for stochastic optimization}. In \bibinfo{booktitle}{\emph{ICLR}}.
\newblock


\bibitem[Kipf and Welling(2017)]%
        {kipf2016semi}
\bibfield{author}{\bibinfo{person}{Thomas~N Kipf} {and} \bibinfo{person}{Max Welling}.} \bibinfo{year}{2017}\natexlab{}.
\newblock \showarticletitle{Semi-supervised classification with graph convolutional networks}. In \bibinfo{booktitle}{\emph{ICLR}}.
\newblock


\bibitem[Kumar et~al\mbox{.}(2016)]%
        {kumar2016edge}
\bibfield{author}{\bibinfo{person}{Srijan Kumar}, \bibinfo{person}{Francesca Spezzano}, \bibinfo{person}{VS Subrahmanian}, {and} \bibinfo{person}{Christos Faloutsos}.} \bibinfo{year}{2016}\natexlab{}.
\newblock \showarticletitle{Edge weight prediction in weighted signed networks}. In \bibinfo{booktitle}{\emph{ICDM}}.
\newblock


\bibitem[Kumar et~al\mbox{.}(2019)]%
        {kumar2019predicting}
\bibfield{author}{\bibinfo{person}{Srijan Kumar}, \bibinfo{person}{Xikun Zhang}, {and} \bibinfo{person}{Jure Leskovec}.} \bibinfo{year}{2019}\natexlab{}.
\newblock \showarticletitle{Predicting dynamic embedding trajectory in temporal interaction networks}. In \bibinfo{booktitle}{\emph{KDD}}.
\newblock


\bibitem[Lee et~al\mbox{.}(2024)]%
        {Lee_SLADE_Online_Appendix_2024}
\bibfield{author}{\bibinfo{person}{Jongha Lee}, \bibinfo{person}{Sunwoo Kim}, {and} \bibinfo{person}{Kijung Shin}.} \bibinfo{year}{2024}\natexlab{}.
\newblock \showarticletitle{{Online Appendix for SLADE}}.
\newblock  (\bibinfo{year}{2024}).
\newblock
\urldef\tempurl%
\url{https://github.com/jhsk777/SLADE-Online-Appendix}
\showURL{%
\tempurl}


\bibitem[Leskovec et~al\mbox{.}(2007)]%
        {leskovec2007graph}
\bibfield{author}{\bibinfo{person}{Jure Leskovec}, \bibinfo{person}{Jon Kleinberg}, {and} \bibinfo{person}{Christos Faloutsos}.} \bibinfo{year}{2007}\natexlab{}.
\newblock \showarticletitle{Graph evolution: Densification and shrinking diameters}.
\newblock \bibinfo{journal}{\emph{ACM Transactions on Knowledge Discovery from Data}} \bibinfo{volume}{1}, \bibinfo{number}{1} (\bibinfo{year}{2007}), \bibinfo{pages}{2--es}.
\newblock


\bibitem[Li et~al\mbox{.}(2017)]%
        {li2017radar}
\bibfield{author}{\bibinfo{person}{Jundong Li}, \bibinfo{person}{Harsh Dani}, \bibinfo{person}{Xia Hu}, {and} \bibinfo{person}{Huan Liu}.} \bibinfo{year}{2017}\natexlab{}.
\newblock \showarticletitle{Radar: Residual analysis for anomaly detection in attributed networks.}. In \bibinfo{booktitle}{\emph{IJCAI}}.
\newblock


\bibitem[Liu et~al\mbox{.}(2021a)]%
        {liu2021cola}
\bibfield{author}{\bibinfo{person}{Yixin Liu}, \bibinfo{person}{Zhao Li}, \bibinfo{person}{Shirui Pan}, \bibinfo{person}{Chen Gong}, \bibinfo{person}{Chuan Zhou}, {and} \bibinfo{person}{George Karypis}.} \bibinfo{year}{2021}\natexlab{a}.
\newblock \showarticletitle{Anomaly detection on attributed networks via contrastive self-supervised learning}.
\newblock \bibinfo{journal}{\emph{IEEE Transactions on Neural Networks and Learning Systems}} \bibinfo{volume}{33}, \bibinfo{number}{6} (\bibinfo{year}{2021}), \bibinfo{pages}{2378--2392}.
\newblock


\bibitem[Liu et~al\mbox{.}(2021b)]%
        {liu2021anomaly}
\bibfield{author}{\bibinfo{person}{Yixin Liu}, \bibinfo{person}{Shirui Pan}, \bibinfo{person}{Yu~Guang Wang}, \bibinfo{person}{Fei Xiong}, \bibinfo{person}{Liang Wang}, \bibinfo{person}{Qingfeng Chen}, {and} \bibinfo{person}{Vincent~CS Lee}.} \bibinfo{year}{2021}\natexlab{b}.
\newblock \showarticletitle{Anomaly detection in dynamic graphs via transformer}.
\newblock \bibinfo{journal}{\emph{IEEE Transactions on Knowledge and Data Engineering}} (\bibinfo{year}{2021}).
\newblock


\bibitem[Ma et~al\mbox{.}(2021)]%
        {ma2021comprehensive}
\bibfield{author}{\bibinfo{person}{Xiaoxiao Ma}, \bibinfo{person}{Jia Wu}, \bibinfo{person}{Shan Xue}, \bibinfo{person}{Jian Yang}, \bibinfo{person}{Chuan Zhou}, \bibinfo{person}{Quan~Z Sheng}, \bibinfo{person}{Hui Xiong}, {and} \bibinfo{person}{Leman Akoglu}.} \bibinfo{year}{2021}\natexlab{}.
\newblock \showarticletitle{A comprehensive survey on graph anomaly detection with deep learning}.
\newblock \bibinfo{journal}{\emph{IEEE Transactions on Knowledge and Data Engineering}} (\bibinfo{year}{2021}).
\newblock


\bibitem[Paranjape et~al\mbox{.}(2017)]%
        {paranjape2017motifs}
\bibfield{author}{\bibinfo{person}{Ashwin Paranjape}, \bibinfo{person}{Austin~R Benson}, {and} \bibinfo{person}{Jure Leskovec}.} \bibinfo{year}{2017}\natexlab{}.
\newblock \showarticletitle{Motifs in temporal networks}. In \bibinfo{booktitle}{\emph{WSDM}}.
\newblock


\bibitem[Pourhabibi et~al\mbox{.}(2020)]%
        {pourhabibi2020fraud}
\bibfield{author}{\bibinfo{person}{Tahereh Pourhabibi}, \bibinfo{person}{Kok-Leong Ong}, \bibinfo{person}{Booi~H Kam}, {and} \bibinfo{person}{Yee~Ling Boo}.} \bibinfo{year}{2020}\natexlab{}.
\newblock \showarticletitle{Fraud detection: A systematic literature review of graph-based anomaly detection approaches}.
\newblock \bibinfo{journal}{\emph{Decision Support Systems}}  \bibinfo{volume}{133} (\bibinfo{year}{2020}), \bibinfo{pages}{113303}.
\newblock


\bibitem[Rossi et~al\mbox{.}(2020)]%
        {tgn_icml_grl2020}
\bibfield{author}{\bibinfo{person}{Emanuele Rossi}, \bibinfo{person}{Ben Chamberlain}, \bibinfo{person}{Fabrizio Frasca}, \bibinfo{person}{Davide Eynard}, \bibinfo{person}{Federico Monti}, {and} \bibinfo{person}{Michael Bronstein}.} \bibinfo{year}{2020}\natexlab{}.
\newblock \showarticletitle{Temporal Graph Networks for Deep Learning on Dynamic Graphs}. In \bibinfo{booktitle}{\emph{ICML 2020 Workshop on Graph Representation Learning}}.
\newblock


\bibitem[Rotman and Wolf(2021)]%
        {rotman2021shuffling}
\bibfield{author}{\bibinfo{person}{Michael Rotman} {and} \bibinfo{person}{Lior Wolf}.} \bibinfo{year}{2021}\natexlab{}.
\newblock \showarticletitle{Shuffling recurrent neural networks}. In \bibinfo{booktitle}{\emph{AAAI}}.
\newblock


\bibitem[Sharma et~al\mbox{.}(2022)]%
        {sharma2022representation}
\bibfield{author}{\bibinfo{person}{Kartik Sharma}, \bibinfo{person}{Mohit Raghavendra}, \bibinfo{person}{Yeon~Chang Lee}, \bibinfo{person}{Srijan Kumar}, {et~al\mbox{.}}} \bibinfo{year}{2022}\natexlab{}.
\newblock \showarticletitle{Representation Learning in Continuous-Time Dynamic Signed Networks}.
\newblock \bibinfo{journal}{\emph{arXiv e-prints}} (\bibinfo{year}{2022}), \bibinfo{pages}{arXiv--2207}.
\newblock


\bibitem[Tian et~al\mbox{.}(2023)]%
        {tian2023sad}
\bibfield{author}{\bibinfo{person}{Sheng Tian}, \bibinfo{person}{Jihai Dong}, \bibinfo{person}{Jintang Li}, \bibinfo{person}{Wenlong Zhao}, \bibinfo{person}{Xiaolong Xu}, \bibinfo{person}{Bowen Song}, \bibinfo{person}{Changhua Meng}, \bibinfo{person}{Tianyi Zhang}, \bibinfo{person}{Liang Chen}, {et~al\mbox{.}}} \bibinfo{year}{2023}\natexlab{}.
\newblock \showarticletitle{SAD: Semi-Supervised Anomaly Detection on Dynamic Graphs}. In \bibinfo{booktitle}{\emph{IJCAI}}.
\newblock


\bibitem[Tian et~al\mbox{.}(2021)]%
        {tian2021self}
\bibfield{author}{\bibinfo{person}{Sheng Tian}, \bibinfo{person}{Ruofan Wu}, \bibinfo{person}{Leilei Shi}, \bibinfo{person}{Liang Zhu}, {and} \bibinfo{person}{Tao Xiong}.} \bibinfo{year}{2021}\natexlab{}.
\newblock \showarticletitle{Self-supervised representation learning on dynamic graphs}. In \bibinfo{booktitle}{\emph{CIKM}}.
\newblock


\bibitem[Trivedi et~al\mbox{.}(2019)]%
        {trivedi2019dyrep}
\bibfield{author}{\bibinfo{person}{Rakshit Trivedi}, \bibinfo{person}{Mehrdad Farajtabar}, \bibinfo{person}{Prasenjeet Biswal}, {and} \bibinfo{person}{Hongyuan Zha}.} \bibinfo{year}{2019}\natexlab{}.
\newblock \showarticletitle{Dyrep: Learning representations over dynamic graphs}. In \bibinfo{booktitle}{\emph{ICLR}}.
\newblock


\bibitem[Vaswani et~al\mbox{.}(2017)]%
        {vaswani2017attention}
\bibfield{author}{\bibinfo{person}{Ashish Vaswani}, \bibinfo{person}{Noam Shazeer}, \bibinfo{person}{Niki Parmar}, \bibinfo{person}{Jakob Uszkoreit}, \bibinfo{person}{Llion Jones}, \bibinfo{person}{Aidan~N Gomez}, \bibinfo{person}{{\L}ukasz Kaiser}, {and} \bibinfo{person}{Illia Polosukhin}.} \bibinfo{year}{2017}\natexlab{}.
\newblock \showarticletitle{Attention is all you need}. In \bibinfo{booktitle}{\emph{NeurIPS}}.
\newblock


\bibitem[Velickovic et~al\mbox{.}(2018)]%
        {velickovic2017graph}
\bibfield{author}{\bibinfo{person}{Petar Velickovic}, \bibinfo{person}{Guillem Cucurull}, \bibinfo{person}{Arantxa Casanova}, \bibinfo{person}{Adriana Romero}, \bibinfo{person}{Pietro Lio}, \bibinfo{person}{Yoshua Bengio}, {et~al\mbox{.}}} \bibinfo{year}{2018}\natexlab{}.
\newblock \showarticletitle{Graph attention networks}. In \bibinfo{booktitle}{\emph{ICLR}}.
\newblock


\bibitem[Wang et~al\mbox{.}(2021b)]%
        {wang2021apan}
\bibfield{author}{\bibinfo{person}{Xuhong Wang}, \bibinfo{person}{Ding Lyu}, \bibinfo{person}{Mengjian Li}, \bibinfo{person}{Yang Xia}, \bibinfo{person}{Qi Yang}, \bibinfo{person}{Xinwen Wang}, \bibinfo{person}{Xinguang Wang}, \bibinfo{person}{Ping Cui}, \bibinfo{person}{Yupu Yang}, \bibinfo{person}{Bowen Sun}, {et~al\mbox{.}}} \bibinfo{year}{2021}\natexlab{b}.
\newblock \showarticletitle{Apan: Asynchronous propagation attention network for real-time temporal graph embedding}. In \bibinfo{booktitle}{\emph{SIGMOD}}.
\newblock


\bibitem[Wang et~al\mbox{.}(2021a)]%
        {wang2021adaptive}
\bibfield{author}{\bibinfo{person}{Yiwei Wang}, \bibinfo{person}{Yujun Cai}, \bibinfo{person}{Yuxuan Liang}, \bibinfo{person}{Henghui Ding}, \bibinfo{person}{Changhu Wang}, \bibinfo{person}{Siddharth Bhatia}, {and} \bibinfo{person}{Bryan Hooi}.} \bibinfo{year}{2021}\natexlab{a}.
\newblock \showarticletitle{Adaptive data augmentation on temporal graphs}. In \bibinfo{booktitle}{\emph{NeurIPS}}.
\newblock


\bibitem[Xu et~al\mbox{.}(2020)]%
        {xu2020inductive}
\bibfield{author}{\bibinfo{person}{Da Xu}, \bibinfo{person}{Chuanwei Ruan}, \bibinfo{person}{Evren Korpeoglu}, \bibinfo{person}{Sushant Kumar}, {and} \bibinfo{person}{Kannan Achan}.} \bibinfo{year}{2020}\natexlab{}.
\newblock \showarticletitle{Inductive representation learning on temporal graphs}. In \bibinfo{booktitle}{\emph{ICLR}}.
\newblock


\bibitem[Yu et~al\mbox{.}(2018)]%
        {yu2018netwalk}
\bibfield{author}{\bibinfo{person}{Wenchao Yu}, \bibinfo{person}{Wei Cheng}, \bibinfo{person}{Charu~C Aggarwal}, \bibinfo{person}{Kai Zhang}, \bibinfo{person}{Haifeng Chen}, {and} \bibinfo{person}{Wei Wang}.} \bibinfo{year}{2018}\natexlab{}.
\newblock \showarticletitle{Netwalk: A flexible deep embedding approach for anomaly detection in dynamic networks}. In \bibinfo{booktitle}{\emph{KDD}}.
\newblock


\bibitem[Zheng et~al\mbox{.}(2019)]%
        {zheng2019addgraph}
\bibfield{author}{\bibinfo{person}{Li Zheng}, \bibinfo{person}{Zhenpeng Li}, \bibinfo{person}{Jian Li}, \bibinfo{person}{Zhao Li}, {and} \bibinfo{person}{Jun Gao}.} \bibinfo{year}{2019}\natexlab{}.
\newblock \showarticletitle{AddGraph: Anomaly Detection in Dynamic Graph Using Attention-based Temporal GCN.}. In \bibinfo{booktitle}{\emph{IJCAI}}.
\newblock


\bibitem[Zheng et~al\mbox{.}(2023)]%
        {zheng2023temporal}
\bibfield{author}{\bibinfo{person}{Tongya Zheng}, \bibinfo{person}{Xinchao Wang}, \bibinfo{person}{Zunlei Feng}, \bibinfo{person}{Jie Song}, \bibinfo{person}{Yunzhi Hao}, \bibinfo{person}{Mingli Song}, \bibinfo{person}{Xingen Wang}, \bibinfo{person}{Xinyu Wang}, {and} \bibinfo{person}{Chun Chen}.} \bibinfo{year}{2023}\natexlab{}.
\newblock \showarticletitle{Temporal Aggregation and Propagation Graph Neural Networks for Dynamic Representation}.
\newblock \bibinfo{journal}{\emph{IEEE Transactions on Knowledge and Data Engineering}} (\bibinfo{year}{2023}).
\newblock


\end{thebibliography}

        \appendix 
        \begin{table*}[!t]
    \centering
    \caption{Statistics of datasets used in our experiments. The count of anomalies refers to the number of edges where the dynamic state of the actor node is abnormal, and the ratio represents the proportion of these abnormal edges in relation to the total number of edges.}
    \scalebox{0.9}{
        \begin{tabular}{lcccccc}
            \toprule
             & Wikipedia & Reddit & Bitcoin-alpha & Bitcoin-OTC & Synthetic-Hijack & Synthetic-New \\
            \midrule
                \# Nodes             & 9,227  & 10,984     & 3,783     & 5,881 & 986 & 996  \\
                \# Edges        & 157,474  & 672,447     & 24,186     & 35,592 & 333,200 & 333,200   \\
                \# Features       & 172  & 172     & 0    & 0 & 0 & 0 \\
                \# anomalies    & 217   & 366      & 874      & 2,568 & 3,290 & 3,290        \\
                anomalies ratio    & 0.14\%   & 0.05\%      & 3.61\%      & 7.22\%  & 0.99\%  & 0.99\%         \\   
                anomaly type    & post ban   & edit ban      & unreliable user      & unreliable user & spammer & spammer       \\
            \bottomrule
        \end{tabular}
    }
    \label{tab:dataset}
\end{table*}
\section{APPENDIX: Dataset Details}
\label{sec:app:dataset}

\subsection{Real-world Datasets}
We use 4 real-world datasets for a dynamic anomaly detection task;
two social network datasets (Wikipedia and Reddit~\citep{kumar2019predicting})~\footnote{http://snap.stanford.edu/jodie/\#datasets} and two financial network datasets (Bitcoin-alpha and Bitcoin-OTC~\citep{kumar2016edge})~\footnote{https://snap.stanford.edu/data}.
Basic descriptive statistics of each dataset are provided in Table~\ref{tab:dataset}.
Below, we provide a detailed description of each dataset.

In Wikipedia and Reddit datasets, which are social networks between users, a user's dynamic label at time $t$ indicates the user's state at time $t$. 
Specifically, if a user is banned by administrators at time $t$, the label of the user at time $t$ is marked as \textit{abnormal}.
Otherwise, the user has a label of \textit{normal}.
Note that these labels are inherently given in the original datasets.

Bitcoin-alpha and Bitcoin-OTC datasets are temporal weighted-signed networks between users. 
Here, the weight of each directional edge indicates how much the source user trusts the destination user.
Specifically, for each edge, its weight, which lies between -10 (total distrust) to +10 (total trust) is given, together with the time of the interaction. 
Note that, in our experiments, the edge weights are utilized to create ground-truth dynamic labels, and they are not included in the inputs for the anomaly detection methods.
While the used social networks inherently contain dynamic node states (i.e., banned), such information of an individual user is not provided in this case.
Therefore, we assign a dynamic label to each node according to the weights of the interactions the node is involved in.
Note that since the overall Bitcoin transaction systems are anonymized, it is not reliable to determine the state of a user according to a single interaction.

Hence, to obtain reliable labels, we adopt a two-stage labeling process: 
for each user, first, we define the (1) overall state of the user, 
then define (2) the dynamic state of the user based on the overall state. 
We further describe this process.
\begin{enumerate}[label={(\arabic*):}, leftmargin = *] 
    \item \textbf{Defining overall state:} For each user, if the sum of the scores a user has received throughout the entire edge stream is smaller than zero, the overall state of the user is decided to be \textit{abnormal}. 
    Otherwise, the overall state of the user is labeled as \textit{normal}.
    \item \textbf{Defining dynamic state:} For each user that has the \textit{normal} overall state, the dynamic label of the user is assumed to be always \textit{normal}.
    Conversely, for each user that has the \textit{abnormal} overall state, when the user becomes the destination node of a negative-weighted interaction (i.e., receiving a negative trust score from other users), the dynamic label of the user at the corresponding interaction time is assumed to be \textit{abnormal}. Otherwise, if a user receives a positive trust score from another user, we assume the dynamic label of the user at that time is assumed to be \textit{normal}.
    Note that these created dynamic node labels are utilized as ground-truth dynamic node labels. 
\end{enumerate}

As mentioned above, the weight of each edge determines the label of a destination node, not a source node\footnotemark. 
Regarding node and edge features, all of the used baseline neural network-based anomaly detection models require feature information of nodes and edges. 
On the other hand, in Bitcoin-alpha and -OTC datasets, such information is not given.
Following prior works~\citep{khoshraftar2022temporal,sharma2022representation} that used such models in the corresponding datasets, we utilize zero vectors for node and edge features of such datasets.

Regarding data splits, for all the datasets, we fix the chronological split with 70 \% for training, 15\% for validation, and the last 15 \% for testing, which is a common setting that is widely used in many existing works~\citep{xu2020inductive,tgn_icml_grl2020, tian2023sad}.

\subsection{Semi-Synthetic Datasets}
Next, we describe the details of the Synthetic-Hijack and Synthetic-New datasets, which we utilize in Section~\ref{sec:exp:email}.
{These datasets are variants of the real-world email communication dataset: Email-EU~\citep{leskovec2007graph}.
Nodes indicate users, and an edge $(v_{i},v_{j},t)$ indicates an email between users $v_{i}$ and $v_{j}$ at time $t$.
Since the dataset does not have ground-truth anomalous interactions, we inject anomalies discussed in Section~\ref{sec:discussion} (\textbf{T1}, \textbf{T2}, and \textbf{T3}).
To this end, we first choose the time interval where interactions have actively occurred (spec., interactions occur within $(4.06\times10^{7},4.54\times10^{7})$ timestamp) since our method and baselines can easily detect anomalies that are injected in the later timestamp than $4.54\times10^{7}$.
Then, we set the timestamp region where the last $10\%$ (in chronicle order) interactions occur as an evaluation region.
After, we select candidate accounts to perform anomalous actions:
\begin{itemize}[leftmargin=*]
    \item \textbf{Synthetic-Hijack:}
    Randomly select 10 nodes ($\approx 1\%$ of the normal nodes) from nodes that are normal before the final 10\% of the dataset and do not appear in the evaluation region. 
    \item \textbf{Synthetic-New:}
    Make 10 new nodes ($\approx 1\%$ of the normal nodes) that have never appeared before the evaluation region. 
\end{itemize}
Subsequently, we inject anomalies equivalent to approximately 1\% of the total normal interactions into the evaluation region (see the next paragraph for details).
Each of these anomalous actions is represented as temporal edge $(v_{k},v_{l},t_{m})$, where at time $t_{m}$, the anomalous source node $v_{k}$ from the selected candidates sends a spam email to some normal node $v_{l}$. 
In this case, the dynamic label of the node $v_{k}$ at time $t_{m}$ is anomalous.

We further elaborate on how we make each anomalous edge. 
We consider the characteristics of spammers. 
We assume that spammers tend to target an unspecified majority with multiple spam emails in a short time interval.
Based on this assumption, (1) we sample a timestamp $t$ in the evaluation region uniformly at random, (2) pick one anomalous node from the candidate pool, regarding it as the source node, (3) randomly choose 10 normal nodes as destinations, and (4) make 10 edges by joining the selected source node with each selected destination nodes.
(5) Lastly, we assign each edge a timestamp $t \pm \alpha$, where $\alpha$ is sampled uniformly at random from $[0, 300]$.
This process ((1) - (5)) is repeated until we have approximately 1\% of anomalies relative to the total normal interactions for the Synthetic-Hijack and Synthetic-New.

\footnotetext{For the sake of consistency with other datasets, the directions are reversed in the Bitcoin datasets, enabling the scoring and evaluation to be performed for source nodes.}
        \begin{table*}[!t]
    \setlength{\tabcolsep}{5pt}
    \setlength\aboverulesep{0.5pt}
    \centering
    \caption{Hyperparameter settings for all baseline methods of main experiments and \method on each dataset.
    }
    \small
    \scalebox{0.9}{
        \begin{tabular}{c|cccc}
            \toprule
            Method & Wikipedia & Reddit & Bitcoin-alpha & Bitcoin-OTC  \\
            \midrule
            \midrule
                \begin{tabular}{@{}c@{}}\SedanSpot \\ \small (sample size, \# random walkers, restart probability)\end{tabular}          & $(5000,200,0.9)$   & $(20000,100,0.9)$  & $(20000,100,0.9)$        & $(5000,200,0.5)$     \\
                \rule{0pt}{15pt}
                \begin{tabular}{@{}c@{}}\MIDAS \\ \small (number of hash functions, CMS size, decay factor)\end{tabular}  & $(3,1024,0.1)$   & $(3,1024,0.5)$  & $(4,256,0.9)$        & $(2,256,0.9)$    \\
                \rule{0pt}{15pt}
                \begin{tabular}{@{}c@{}}\FFADE \\ \small (batch size, memory size, embedding size)\end{tabular}  & $(100,100,100)$   & $(300,200,200)$  & $(100,400,100)$        & $(100,100,100)$      \\
                \rule{0pt}{15pt}
                \begin{tabular}{@{}c@{}}\Anoedgel \\ \small (edge threshold, bucket size, decay factor) \end{tabular}            & $(50,128,0.5)$  & $(50,64,0.1)$  & $(50,128,0.9)$        & $(50,128,0.5)$    \\
            \midrule
                \begin{tabular}{@{}c@{}}\JODIE \\ \small (pretraining batch size, batch size, node degree)\end{tabular}    & $(300,300,20)$   & $(200,100,10)$   & $(200,100,10)$  & $(200,100,10)$   \\\rule{0pt}{15pt}
                \begin{tabular}{@{}c@{}}\Dyrep \\ \small (pretraining batch size, batch size, node degree)\end{tabular} & $(200,200,10)$   & $(200,100,20)$   & $(200,100,10)$   & $(100,300,20)$   \\\rule{0pt}{15pt}
                \begin{tabular}{@{}c@{}}\TGAT \\ \small (pretraining batch size, batch size, node degree)\end{tabular} & $(200,100,20)$   & $(200,100,20)$   & $(200,100,20)$   & $(300,300,10)$   \\\rule{0pt}{15pt}
                \begin{tabular}{@{}c@{}}\TGN \\ \small (pretraining batch size, batch size, node degree)\end{tabular} & $(200,100,20)$   & $(200,100,10)$   & $(200,100,10)$   & $(200,100,10)$     \\\rule{0pt}{15pt}
                \begin{tabular}{@{}c@{}}\SAD \\ \small (batch size, anomaly alpha, supervised alpha)\end{tabular}   & $(100,0.1,0.01)$     &$(256,0.1,0.0005)$   & $(300,0.001,0.001)$  & $(200,0.01,0.1)$        \\
            \midrule
                \begin{tabular}{@{}c@{}}\method \\ \small (batch size, $\omega_{gs}$, $\omega_{gd}$)\end{tabular}   & $(100,0.1,0.1)$     & $(100,0.1,0.1)$       & $(100,0.1,0.1)$       & $(100,0.1,0.1)$        \\\rule{0pt}{15pt}
                \begin{tabular}{@{}c@{}}\method-HP \\ \small (batch size,$ \omega_{gs}$, $\omega_{gd}$)\end{tabular}   & $(300,10,10)$     &$(100,0.1,1)$       & $(300,1,10)$       & $(300,10,1)$        \\
            \bottomrule
        \end{tabular}
        }
    \normalsize
    \label{tab:HP_setting}
\end{table*}
\section{Appendix: Implementation Details}
\label{sec:app:impl}


\subsection{Experiments Environment}
\label{sec:app:impl:infra}
We conduct all experiments with NVIDIA RTX 3090 Ti GPUs (24GB VRAM), 256GB of RAM, and two Intel Xeon Silver 4210R Processors.

\subsection{Details of Baselines in Main Experiments}
\label{sec:app:impl:baseline_param}

As mentioned in Section~\ref{sec:exp:details}, we tune most hyperparameters of each baseline method by conducting a full grid search on the validation set of each dataset.
For other hyperparameters, we strictly follow the setting provided in their official code, because it leads to better results than grid searches.
The selected hyperparameter combination of each model is reported in Table~\ref{tab:HP_setting}.
\paragraph{Rule-based Methods} Hyperparameter search space of each rule-based method is as below:
\begin{itemize}[leftmargin=*]
    \item \textbf{Sedanspot}: sample size among $(5000,10000,20000)$, number of random walkers among $(100,200,300)$, and restart probability among $(0.1,0.5,0.9)$
    \item \textbf{MIDAS-R}: number of hash functions among $(2,3,4)$, CMS size among $(256,512,1024)$, and decay factor among $(0.1,0.5,0.9)$
    \item \textbf{F-FADE}: batch size among $(100,200,300)$, limited memory size among $(100,200,400)$, and embedding size among $(100,200,300)$
    \item \textbf{Anoedge-l}: edge threshold among $(25,50,100)$, bucket size among $(64,128,256)$, and restart probability among $(0.1,0.5,0.9)$
\end{itemize}

\paragraph{Neural Network-based Methods} We train all models with the Adam ~\citep{kingma2014adam} optimizer. 
The hyperparameter search space of each neural network-based method is as below:
\begin{itemize}[leftmargin=*]
    \item \textbf{JODIE}: self-supervised batch size among (100, 200, 300), supervised batch size among $(100, 200, 300)$, and degree among $(10, 20)$
    \item \textbf{Dyrep}: self-supervised batch size among (100, 200, 300), supervised batch size among $(100, 200, 300)$, and degree among $(10, 20)$
    \item \textbf{TGAT}: self-supervised batch size among (100, 200, 300), supervised batch size among $(100, 200, 300)$, and degree among $(10, 20)$
    \item \textbf{TGN}: self-supervised batch size among (100, 200, 300), supervised batch size among $(100, 200, 300)$, and degree among $(10, 20)$
    \item \textbf{SAD}: batch size among $(100,200,300)$, anomaly alpha among $(0.1,0.01,0.001)$ for deviation loss, and supervised alpha among $(0.1,0.01,0.001)$ for supervised loss    
\end{itemize}
 
\begin{table}[!t]
    \setlength{\tabcolsep}{5pt}
    \setlength\aboverulesep{0.5pt}
    \centering
    \caption{Hyperparameter settings for rule-based baseline methods and \method in type analysis.
    }
    \small
    \scalebox{0.9}{
        \begin{tabular}{c|c}
            \toprule
            Method & Synthetic-$(*)$  \\
            \midrule
            \midrule
                \begin{tabular}{@{}c@{}}\SedanSpot \\ \small (sample size, \# random walkers, restart probability)\end{tabular}          & $(10000,50,0.15)$  \\
                \rule{0pt}{15pt}
                \begin{tabular}{@{}c@{}}\MIDAS \\ \small (number of hash functions, CMS size, decay factor)\end{tabular}  & $(2,1024,0.5)$      \\
                \rule{0pt}{15pt}
                \begin{tabular}{@{}c@{}}\FFADE \\ \small (batch size, memory size, embedding size)\end{tabular}  & $(100,100,100)$     \\
                \rule{0pt}{15pt}
                \begin{tabular}{@{}c@{}}\Anoedgel \\ \small (edge threshold, bucket size, decay factor) \end{tabular}            & $(50,32,0.9)$   \\
            \midrule
                \begin{tabular}{@{}c@{}}\method \\ \small (batch size, $\omega_{gs}$, $\omega_{gd}$)\end{tabular}   & $(100,0.1,0.1)$       \\
            \bottomrule
        \end{tabular}
        }
    \normalsize
    \label{tab:HP_setting_real}
\end{table}
\subsection{Details of Baseline Methods in Type Analysis Experiments}
\label{sec:app:impl:baseline_param_real}
 We use hyperparameter settings from previous studies or combinations for good performance on our main experiments.
The selected hyperparameter setting of each model is reported in Table
~\ref{tab:HP_setting_real}.

\subsection{Detailed Hyperparameters of \method and \method-HP}
\label{sec:app:impl:model_param}
We train both \method and \method-HP using the Adam optimizer, with a learning rate of $3\times10^{-6}$ and a weight decay of $10^{-4}$. 
We fix the dropout probability and the number of attention heads in TGAT to 0.1 and 2, respectively.
In addition, we fix the scaling scalar of temporal encoding (Eq (2) in the main paper) to $\alpha = 10, \beta = 25.6$, the weight of each loss in $\calL_{c}$ (Eq (7) in the main paper) to $\omega_{cs} = \omega_{cd} = 1$, and the dimension of a time encoding to 256, which is equivalent to the memory dimension.
As mentioned in Section 6.1 of the main paper, we tune several hyperparameters of \method-HP on the validation set of each dataset, as we tune the hyperparameters of all baselines. The search space is as follows:
\begin{itemize}[leftmargin=*]
    \item \textbf{\method-HP}: batch size among $(100, 200, 300)$, the weight in memory generation loss $\calL_{g}$ for a source node ($\omega_{gc}$) among $(0.1,1,10)$, and for a destination node ($\omega_{gd}$) among $(0.1,1,10)$.
\end{itemize}
Our final hyperparameter settings for \method and \method-HP are also reported in Table~\ref{tab:HP_setting}.

\end{document}